\documentclass[sigconf,anonymous=False]{acmart}

\copyrightyear{2022}
\acmYear{2022}
\setcopyright{acmlicensed}\acmConference[CIKM '22]{Proceedings of the 31st ACM International Conference on Information and Knowledge Management}{October 17--21, 2022}{Atlanta, GA, USA}
\acmBooktitle{Proceedings of the 31st ACM International Conference on Information and Knowledge Management (CIKM '22), October 17--21, 2022, Atlanta, GA, USA}
\acmPrice{15.00}
\acmDOI{10.1145/3511808.3557460}
\acmISBN{978-1-4503-9236-5/22/10}

\usepackage[framemethod=TikZ]{mdframed}
\usepackage{wrapfig}
\usepackage{amsmath,amsfonts}
\usepackage{dsfont}
\usepackage{fancyvrb}
\usepackage{graphicx}
\usepackage{multirow}
\usepackage{booktabs}
\usepackage{balance}
\usepackage{subcaption}
\usepackage{setspace}
\usepackage{pifont}
\usepackage{xcolor}
\usepackage{graphics}
\usepackage{svg}
\usepackage{dblfloatfix}
\usepackage{float}
\usepackage{circledsteps}
\usepackage{xspace}
\usepackage[ruled,vlined]{algorithm2e}
\usepackage{algorithmic}
\SetKwComment{Comment}{\# }{}
\usepackage{xspace}

\graphicspath{{images/}}

\newcommand{\peq}{\mathrel{{+}{=}}}

\newcommand{\method}{\textsc{Stop}\&\textsc{Hop}\xspace}

\definecolor{tblue}{RGB}{93, 142, 150}
\definecolor{tred}{RGB}{191, 97, 106}
\definecolor{dlblue}{RGB}{216, 235, 255}
\definecolor{dgreen}{RGB}{124, 155, 127}
\definecolor{dpink}{RGB}{207, 166, 208}
\definecolor{dyellow}{RGB}{255, 248, 199}
\definecolor{dgray}{RGB}{46, 49, 49}

\definecolor{cred}{RGB}{191, 97, 106}

\newcommand{\durl}[1]{\textcolor{tblue}{\underline{\url{#1}}}}

\newcommand*\circled[1]{\tikz[baseline=(char.base)]{\node[shape=circle,draw,inner sep=0.7pt] (char) {\footnotesize{#1}};}}
\newmdenv[
  topline=false,
  bottomline=false,
  rightline = false,
  leftmargin=10pt,
  rightmargin=0pt,
  innertopmargin=0pt,
  innerbottommargin=0pt
]{innerproof}

\newcounter{DaveDefCounter}
\begin{document}

% \title{Learning to Stop Early and Classify Ongoing Irregular Time Series}
\title[\textsc{Stop}\&\textsc{Hop}: Early Classification of Irregular Time Series]{\textsc{Stop}\&\textsc{Hop}: Early Classification of Irregular Time Series}

\author{Thomas Hartvigsen}
\affiliation{
  \institution{Massachusetts Institute of Technology}
  \country{}
}
\email{tomh@mit.edu}

\author{Walter Gerych}
\affiliation{
  \institution{Worcester Polytechnic Institute}
  \country{}
}
\email{wgerych@wpi.edu}

\author{Jidapa Thadajarassiri}
\affiliation{
  \institution{Worcester Polytechnic Institute}
  \country{}
}
\email{jthadajarassiri@wpi.edu}

\author{Xiangnan Kong}
\affiliation{
  \institution{Worcester Polytechnic Institute}
  \country{}
}
\email{xkong@wpi.edu}

\author{Elke Rundensteiner}
\affiliation{
  \institution{Worcester Polytechnic Institute}
  \country{}
}
\email{rundenst@wpi.edu}

\renewcommand{\shortauthors}{T. Hartvigsen et al.}

\begin{abstract}
Early classification algorithms help users react faster to their machine learning model's predictions. Early warning systems in hospitals, for example, let clinicians improve their patients' outcomes by accurately predicting infections. While early classification systems are advancing rapidly, a major gap remains: existing systems do not consider irregular time series, which have \textit{uneven} and \textit{often-long} gaps between their observations. Such series are notoriously pervasive in impactful domains like healthcare. We bridge this gap and study early classification of irregular time series, a new setting for early classifiers that opens doors to more real-world problems. Our solution, \method, uses a continuous-time recurrent network to model ongoing irregular time series in real time, while an irregularity-aware halting policy, trained with reinforcement learning, predicts when to stop and classify the streaming series. By taking real-valued step sizes, the halting policy flexibly decides exactly when to stop ongoing series in real time. This way, \method seamlessly integrates information contained in the \textit{timing of observations}, a new and vital source for early classification in this setting, with the time series values to provide early classifications for irregular time series. Using four synthetic and three real-world datasets, we demonstrate that \method consistently makes earlier and more-accurate predictions than state-of-the-art alternatives adapted to this new problem. Our code is publicly available at \texttt{https://github.com/thartvigsen/StopAndHop}.

\end{abstract}



\maketitle

\section{Introduction}
\label{sec:intro}
\begin{figure}[t]
    \centering
    \includegraphics[width=\linewidth]{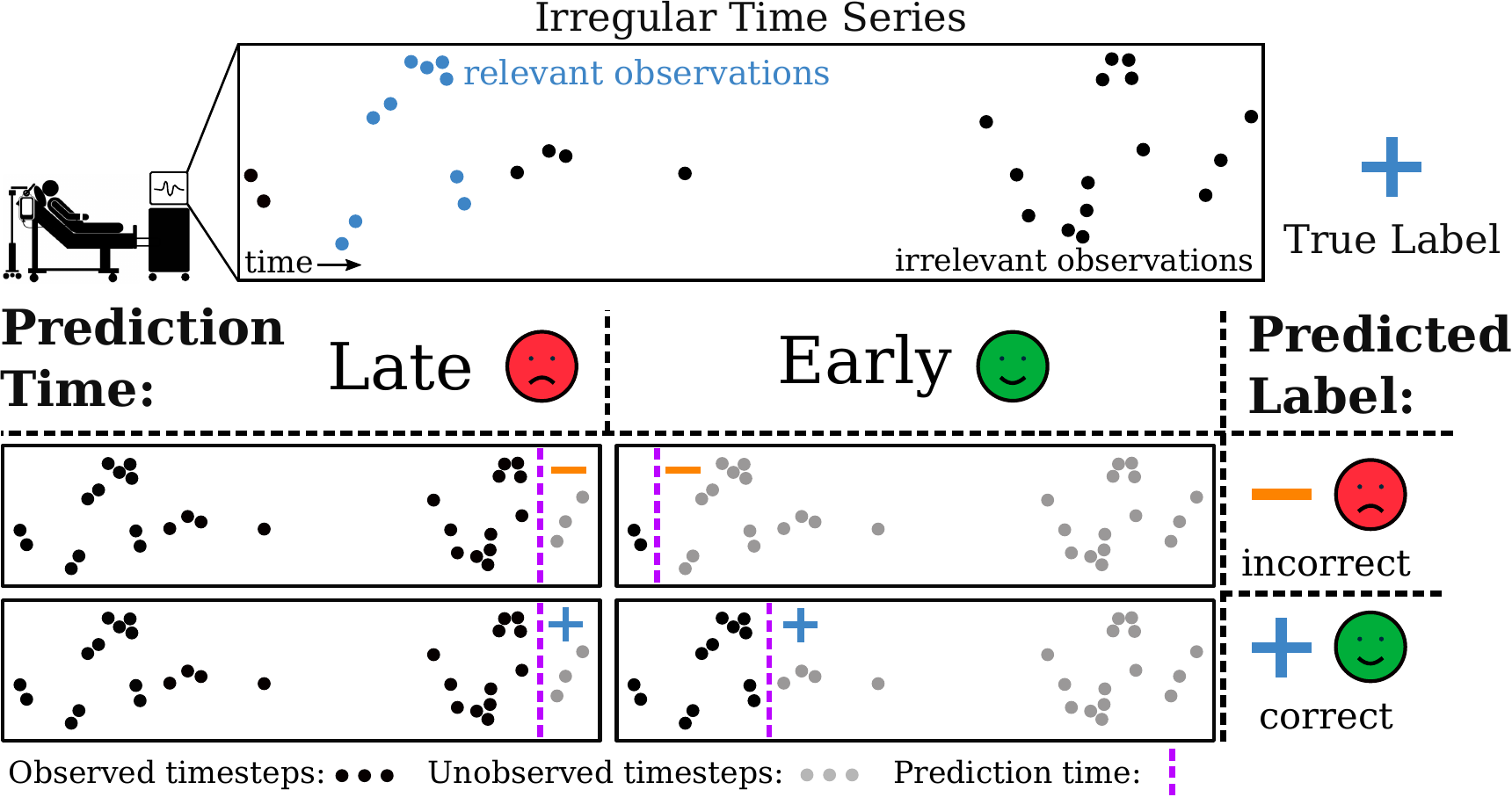}
    \caption{Early Classification of Irregular Time Series. Purple vertical dashed lines indicate the prediction time, disregarding all future data. The goal is to classify irregular series while using as little of the timeline as possible.}
    \label{fig:prob_def}
\end{figure}

\textbf{Background.}
Early Classification algorithms classify time series before observing every timestep \cite{xing2009early}.
These powerful methods give users enough time to react in time-sensitive domains.
% Time-sensitive domains require these algorithms when users need ample time to react.
For example, early warning systems help clinicians improve their patients' outcomes \cite{yala2022optimizing,zhao2019asynchronous}.
Still, the best time to classify depends on the task and a given time series because signals are often sporadic.
Too early and it may be a false alarm, fatiguing the end-user \cite{graham2010monitor}.
Too late and the user cannot react.
An ideal classification is both early and accurate, but these goals conflict: an early classification may ignore relevant future data, as illustrated in Figure \ref{fig:prob_def}.

\textbf{State-of-the-art.}
% Trim version
Early classification of time series (ECTS) is advancing quickly.
Recent approaches overcome the poor scaling and false alarms inherent to classic ECTS \cite{wu2021early} via reinforcement learning, predicting whether to Stop or Wait at every step of an ongoing series \cite{martinez2019adaptive,hartvigsen2019adaptive,hartvigsen2020recurrent,yala2022optimizing}.
This approach counteracts overconfident classifiers, resulting in better predictions.
In addition, by using neural networks they classify multivariate time series seamlessly, a setting known to challenge similarity search methods \cite{ghalwash2013extraction,ghalwash2014utilizing,he2015early,ye2009time,xing2011extracting}.
However, the state-of-the-art methods, along with all prior ECTS algorithms, disregard uneven gaps between observations. %; large gaps amplify their waiting costs.

\begin{figure}[t]
    \centering
    \includegraphics[width=0.95\linewidth]{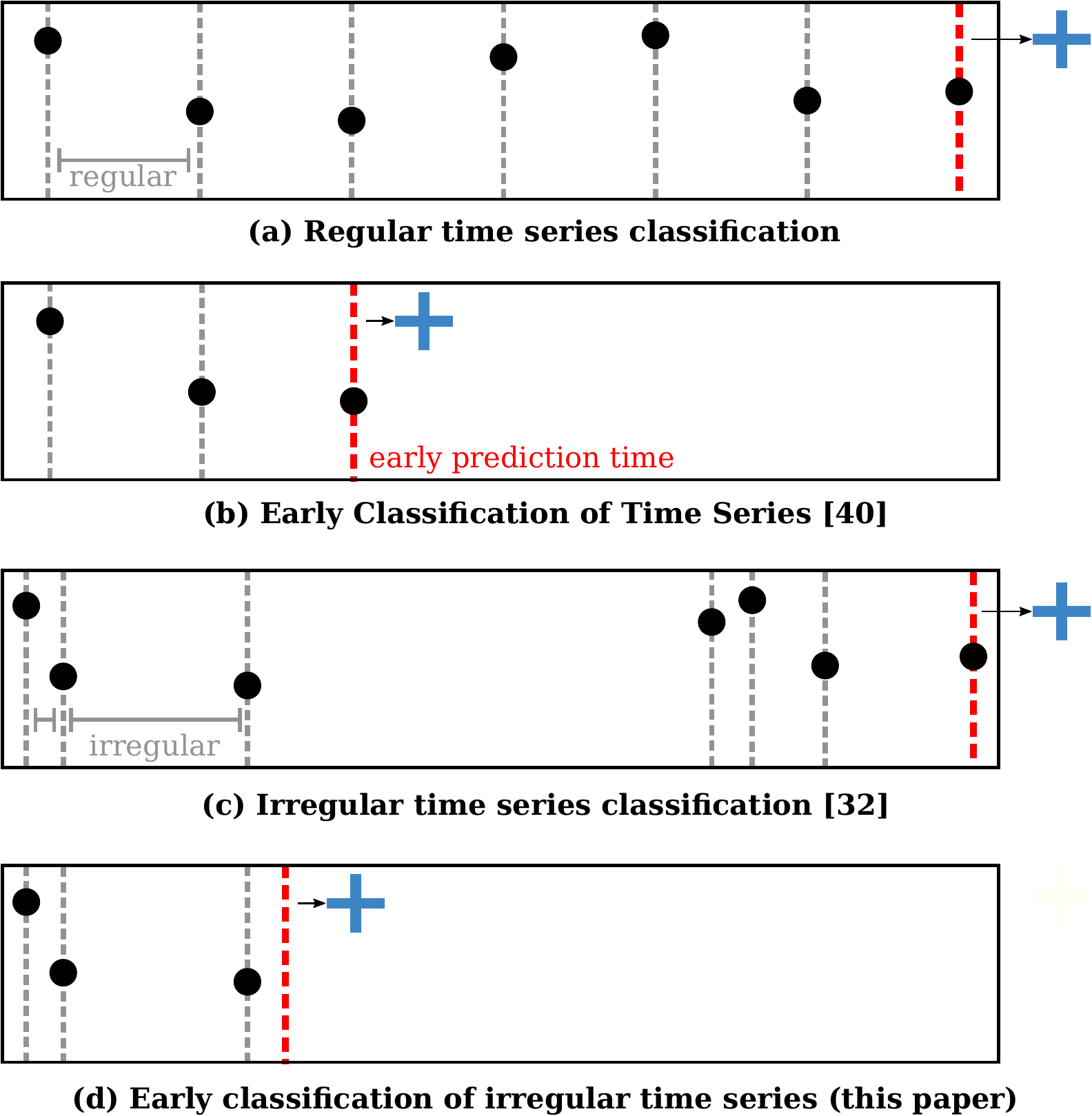}
    \caption{Comparing our problem, early classification of irregular time series (d) to other related problems. (a) Shows standard time series classification, using all possible observations.
    (b) Shows standard early classification of time series, requiring evenly-spaced observations.
    (c) Shows standard irregular time series classification of irregular time series, using all possible observations.
    % Each problem has an analogous multivariate setting.
    % \wnote{My eyes focused first on the dark points instead of the gray, as the dark s easier to see. Unfortunately, the dark points in (b) look like they have irregular gaps between them just because the change in the $y$ direction is so much (so they kind of look irregular because the gap between the 5th and 6th point looks so big, for instance, even though they are even though i can see they're evenly samples when i look closer). so it took me a second to see the different between (b) and (c). }
    }
    \label{fig:family_portrait}
    % \vspace{-5mm}
\end{figure}

\textbf{Knowledge Gap.}
Despite rapid improvements, there remain ample opportunities to broaden the reach of early classifiers.
In particular, existing early classifiers require that their input time series have even spaces between observations: they decide whether to stop early whenever any new observation arrives.
% However, irregular time series---having uneven and possibly-large gaps between their observations---are extremely common in practice, appearing in domains such as healthcare \cite{lipton2016directly}, climate science \cite{bahadori2012granger}, and astronomy \cite{richards2011machine}.
% In a hospital, for instance, knowledge of a patient's health largely comes from doctors measuring a set of variables (\textit{e.g.}, test results and vital sign recordings) over time.
% % Within each variable, there are often irregular gaps between observations.
% Further, irregular time series are often multivariate and irregular between variables as well.
% When faced with irregular time series, existing methods fail because they simply wait for the next observation to arrive before making a decision. This may be arbitrarily-far into the future and misses an opportunity to learn from the \textit{timing} of observations \cite{lipton2016directly}.
% However, in many time-sensitive domains, time series observations arrive
% \textit{sporadically}, only recording \textit{partial} information about an ongoing system.
% Such Irregular Time Series (ITS) thus have uneven gaps between their observations and are common to many
% \wnote{the text from here to the rest of the paragraph is mostly repeat of the above; i do like the second way of wording it better, is that the one you're thinking o fgoing with?}
Meanwhile, irregular time series (ITS), which have \textit{uneven} and \textit{often-large gaps} between observations, are ubiquitous in impactful domains like healthcare \cite{lipton2016directly}, climate science \cite{bahadori2012granger}, and astronomy \cite{richards2011machine}.
In a hospital, for example, clinicians quantify their patient's health over time by taking measurements (laboratory tests, vital signs, microbiology tests, etc.).
% Within each variable, there are often irregular gaps between observations.
Such ITS are often multivariate and irregular between variables: measuring one variable does not mean measuring \textit{all}. % just because one variable is measured does not mean all others are.

To model ITS data effectively, we must acknowledge and leverage asynchronous observations \cite{rubanova2019latent}.
Specifically, successful ITS classifiers should consider \textit{patterns of irregularity}; different classes often have unique signatures in their observations' timing \cite{lipton2016directly}.
For example, diabetes-risk patients receive more blood sugar tests while in the hospital.
The early classification community has yet to consider this common, rich, and general type of time series data.
% Existing early classification methods already struggle with multivariate series \cite{he2015early}, and many are inapplicable without large assumptions in this case.
% \wnote{think part of this paragraph is meant to be commented out - we're repeating ourselves}
% Further, multiple variables are rarely recorded at the exact same time, creating irregularity between variables.
% Learning to classify ITS \textit{early} is an impactful and unexplored problem that can improve outcomes in time-sensitive domains.

% --- Problem Definition and Challenges ---
\textbf{Problem Definition.}
Our work is the first to consider early classification of irregular time series, which is an open, impactful, and challenging real-world problem.
For a previously-unseen ITS $X$, we seek one small (early) real-valued time $\tau$ at which the entire series $X$ may be classified accurately without using any observations later than $\tau$.
The goals of earliness and accuracy naturally conflict because early predictions are usually made at the expense of accuracy (fewer observations have been collected).
This is a multi-objective problem, which we compare to other settings in Figure \ref{fig:family_portrait}.
A key application of this problem is in healthcare, where clinicians require early and accurate predictions, yet use almost entirely with irregular time series \cite{jane2016temporal,yala2022optimizing,zhao2019asynchronous,alavi2022real}.

\textbf{Challenges.}
Our problem is challenging for three main reasons:
\begin{itemize}
    \item \textit{Conflicting Objectives}: Earliness and accuracy contradict each other, so a balance must be struck, where the relative importance of each goal may change depending on the task.
    This trade-off may also change for each instance. % \tnote{better?}
    % according to both the task at hand and each input instance. \wnote{a balance between the task and the instance iisn't 100\% clear to me - maybe ``...a balance must be struck between performance and earliness, where the relative importance can change for each instance"? or ``...a balance must be struck between performance and earliness for each instance"? }
    \item \textit{Unknown Halting Times}: The optimal halting time $\tau$ is unobserved and unavailable for supervision or evaluation.
    \item \textit{Irregular, multivariate observations}: ITS are often sparse and multivariate. This is notoriously challenging for most machine learning methods, which typically require fixed-length and regularly spaced inputs. Signals for earliness and accuracy can also be found in patterns of missingness.
\end{itemize}

\textbf{Proposed Method.}
We overcome these challenges, proposing the first early classifier for irregular time series, which we refer to as \method in reference to our key idea.
\method instantiates a general and modular solution, integrating three essential components.
First, a continuous-time recurrent \textit{Prefix Encoder} embeds ITS data up to a candidate \textit{halting time}, expanding on recent successes in representation learning for ITS.
This handles irregular, multivariate observations by producing dense vector representations of ongoing time series.
Then, a reinforcement learning-based \textit{Halting Policy Network} decides whether or not to \textit{Stop} and predict or \textit{Wait} for more data. If it chooses to Wait, it predicts \textit{for how long} and \textit{Hops} forward in time before reactivating.
Thus we formulate the early classification problem as a Partially-Observable Markov Decision Process with actions that operate on varying time scales.
This formulation lets us encourage early stopping, even though the true halting times are unknown.
The policy can thus adapt to variations in the irregularity of each input instance, using the observations' timing to inform the halting time.
Once the Halting Policy Network stops, a \textit{Prefix Classifier Network} classifies the series.
All components are trained jointly so that the \textit{conflicting objectives}, earliness and accuracy, are easily balanced.

Advancing beyond prior methods, \method \circled{1} accesses more possible prediction times, and \circled{2} learns directly from observation timing, a new source of information for early classification systems.
% \method improves over existing methods by \circOne accessing a larger space of possible prediction times.% considering more times to make predictions
% ---the Halting Policy Network predicts continuous-valued step-sizes (prior methods take hand-picked discrete steps), and \circTwo learning from \textit{observation timing}, a new source of information for early classification systems.
These improvements make a new state-of-the-art early classifier.

\textbf{Contributions.}
Our contributions are as follows:
\begin{enumerate}
    \item We define the open problem \textit{Early Classification of Irregular Time Series}, bridging a major gap between modern early classifiers and real time-sensitive decision making.
    \item Our method, \method, is the first solution to this problem, integrating a continuous-time representation learner with two cooperative reinforcement learning agents.
    \item We show that \method classifies ITS earlier and more-accurately than state-of-the-art alternatives. We also show that \method learns to stop \textit{exactly} when signals arrive using four synthetic datasets, leading to the earliest-possible classifications.
    % learn when to stop on three real-world time-sensitive tasks, outperforming state-of-the-art alternatives. \wnote{can we say something like ``We also show that \method makes predictions earlier and more accurately than the state-of-the-art" or something? might be stronger than ``learns to stop", since we don't/can't say ``learns to stop as early as possible" for the real-world datasets (since we don't know the earliest point)}
\end{enumerate}

\section{Related Work}
\label{sec:related_works}
% MAKE LIMITATIONS OF TRANSFERRING TO MISSING VALUES PROBLEM CLEAR
\textbf{Early Classification of Time Series.}
Early Classification of Time Series (ECTS) is a machine learning problem: correctly predict the label of a streaming time series using as few observations as possible \cite{gupta2020approaches}.
% The goal is to correctly predict the label of a streaming time series using as few observations as possible. %before waiting until it has been observed fully.
Solutions choose one early timestep per time series at which the whole instance is classified (without cheating and looking at future values).
% One early timestep is chosen per time series at which the whole instance is classified.
Classifying sequences early is classically targeted at time series
\cite{he2015early,xing2011extracting,xing2012early,ghalwash2014utilizing,ghalwash2013extraction,mori2018early,schafer2020teaser}.
% some works have also extended to text \cite{huang2017length} and video \cite{ma2016learning}.
Most recent approaches \cite{hartvigsen2020recurrent,hartvigsen2019adaptive,martinez2019adaptive,dennis2018multiple,ebihara2020sequential} have turned to deep learning, extending beyond traditional methods for univariate time series \cite{mori2017reliable,xing2012early,xing2009early,xing2011extracting}, which scale poorly by exhaustively searching for discriminative subsequences \cite{he2015early}.
The current best solution is to frame this problem as a Partially-Observe Markov Decision Process, where at each regularly-spaced timestep, a policy decides whether or not to stop and predict the label. % \cite{martinez2019adaptive,hartvigsen2019adaptive,hartvigsen2020recurrent}.
Some halt RNNs early \cite{hartvigsen2019adaptive,hartvigsen2020recurrent} while others use Deep Q-Networks \cite{martinez2019adaptive}.
%Despite not solving an MDP, more classic ECTS methods also assume regularly-spaced inputs, deciding whether or not to halt depending on distance metric thresholds on shapelets \cite{ye2009time}.
% \wnote{``also assume" - not clear that the MDP methods assume evenly spaced until now. at least to someone not familiar with MDPs.}

\begin{figure*}[htp]
    \centering
    \includegraphics[width=\textwidth]{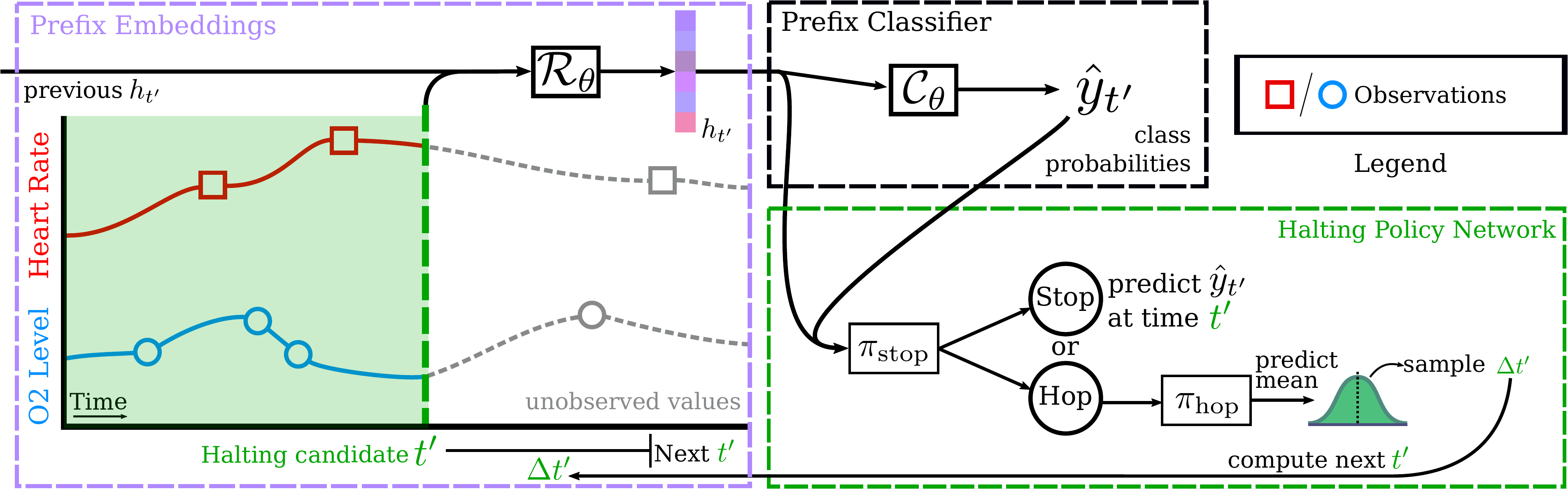}
    \caption{\method Architecture. Given a timestamp $t^\prime$, an embedding is computed for all values and irregularity prior to $t^\prime$. Then, a classifier attempts to classify the series. The halting policy network then uses this classification and the embedding to decide whether or not to stop or, if not, how long to wait before repeating this process.}
    \label{fig:architecture}
\end{figure*}

A major limitation of existing ECTS methods is their reliance on inputs being regularly-spaced; they decide whether or not to halt at each possible timestep.
This does not account for missing values or gaps between observations,
features essential to classifying ITS \cite{che2018recurrent}. 
In ITS, the gaps between consecutive observations may even be large and
unpredictable, so waiting until the next value arrives has consequences. % may be large gaps between consecutive observations and so waiting until the next timestep may add a huge delay to prediction.
Furthermore, many ITS are multivariate, and multiple variables are rarely observed concurrently.
This compounds issues with existing ECTS methods.
Further, \textit{the times at which observations arrive} can itself provide valuable knowledge for accuracy  \cite{lipton2016directly} and earliness.
A successful solution to our problem should take advantage of this extra source of information.

\textbf{Learning from Irregular Time Series.}
Standard machine learning techniques often fail for ITS as they assume fixed-length and regularly-spaced inputs \cite{shukla2020survey}, which are especially rare in important medical settings \cite{sun2020review}.
To bridge this gap, myriad recent works learn from ITS \textit{directly}, developing models that take irregular series as inputs.
Some approaches augment RNNs by either including auxiliary information such as a \textit{missingness-indicator} \cite{lipton2016directly} or \textit{time-since-last-observation} \cite{che2018recurrent} as extra features to preserve properties found in the irregularity.
Others build more complex value estimators by either learning generative models \cite{cheng2020learning}, using gaussian kernel adapters \cite{shukla2019interpolation,li2016scalable}, set functions \cite{horn2020set}, or including decay mechanisms in Recurrent Neural Networks (RNN) to encode information-loss when variables go unobserved over long periods of time \cite{mozer2017discrete,che2018recurrent}.
Some recent works have begun parameterizing ordinary differential equations to serve as time series models \cite{kidger2020neural,lechner2020learning,rubanova2019latent,jia2019neural}, 
Some very recent models have also begun to integrate attention mechanisms into this estimation process \cite{shukla2021multi,chen2021continuous,tan2020data}.
%,song2018attend}.

However, ITS model considers \textit{when} to return predictions to end users in the ongoing timeline. %in the continuous timeline of an ongoing series they should return a prediction to the end user.
A key constraint of the early classification of irregular time series problem is that when classifying a series at a particular point in its timeline, \textit{we cannot use any future values}.
This constraint hinders the use of methods that use all observations during interpolation \cite{shukla2019interpolation,shukla2021multi} or ODE models that encode sequences backwards \cite{rubanova2019latent}.
Such methods are quickly becoming pervasive \cite{kidger2020neural}, though creating online models for ITS remains a burgeoning area \cite{morrill2021neural,cheng2020sparse,weerakody2021review}.

\section{Problem Formulation}
\label{sec:methods}
Assume we are given a set of $N$ labeled irregular time series
$\mathcal{D} = \{(X^i, y^i)\}_{i=1}^N$.
Each series $X$ is a collection of one sequence of $T^d$ (timestep, value) pairs per variable $d$: $X = \{\{(t^d_j, v^d_j)\}_{j=1}^{T^d}\}_{d=1}^D$ where each sequence of timesteps is strictly increasing ($t^d_1 < t^d_2 < \dots t^d_{T^d}$) and $v^d_j$ is the corresponding value of variable $d$ for each timestep.
$T^d$ denotes the number of observations for variable $d$.
$X$ is irregular in that typically $t_i^j \neq t_i^k$ and $t_{i+1}^j - t_i^j \neq t_{i+2}^j - t_{i+1}^j$ for all $i$, $j$, and $k$.
Each label $y$ indicates to which of $C$ classes $X$ belongs.
Our goal of Early Classification of Irregular Time Series is to learn a function $f$ that maps previously-unseen input time series to their accurate class labels $y$ based \textit{only} on values observed prior to some early time $\tau$, which is an unknown function of $X$.
% \wnote{should $\tau$ be a function of $X$? not every time series should halt before the same set time; it should change based on the time series.} % \in [T_\text{min}, T_\text{max}$ where $T_\text{min}$ and $T_\text{max}$ are the smallest and largest timestamps in $X$, respectively. % \in [0, T_\text{max}]$, where $T_\text{max}$ is the largest timestep in series $X$. %(\textit{i.e.} $f(X_{\leq\tau}) = \hat{y}$)
The smaller $\tau$ is, the better.
However, fully achieving both goals at the same time in practice is usually impossible since early predictions are often made at the expense of accuracy as less of the series has been observed.
Thus we seek a tunable solution that balances \textit{earliness} and \textit{accuracy} according to the task at hand.

% \wnote{we don't explicitly say $y^i$ is the class, or what assumptions we have on $y^i$ (binary, multi-class, multi-label). Might not matter but maybe better to clearly define the classification setting?}

% \wnote{Should we more clearly say the $\tau$ might not be one of the time steps where we observe a reading? that $\tau$ can be any value between 0 and the max?}

\section{Method}
\label{sec:methods}
% \begin{algorithm}[htp]
% \SetAlgoLined
% \KwResult{$\hat{y}$, $t$}
%  $t=0$\;
%  \While{$t \leq t_{max}$}{
%   $\hat{X}_t = \text{\texttt{Estimator}}(t)$\;
%   $h = \text{\texttt{HaltingPolicy}}(\hat{X}_t)$\;
%   \eIf{$h=1$}{
%   $\hat{y} = \text{\texttt{Classifier}}(\hat{X}_{\leq t})$\;
%   break\;
%   }{
%   $t = \text{\texttt{Locator}}(\hat{X}_{\leq t})$\;
%   }
%  }
%  \caption{Continuous-Time Early Classification}
% \end{algorithm}

% \begin{wrapfigure}{RT}{0.47\textwidth}
% \begin{figure}[t]
% \begin{algorithm}[h]
% \begin{algorithmic}
% \SetAlgoLined
% \KwResult{Class prediction $\hat{y}$, Halting time $\tau$}
%     $t^\prime = 0$\;
%     $a_{t^\prime} = 0$\;
%     $h_{t^\prime} = 0 \in \mathbb{R}^{D}$\;\\
%     \While{$a_{t^\prime} \neq 1$ and $t^\prime \leq T_\text{max}$}{
%         $h_{t^\prime} = \text{ITSPrefixEncoder}(X, t^\prime)$\;\\
%         $a_{t^\prime} = \text{HaltingPolicyNetwork}(h_{t^\prime})$\;\\
%     \If{$a_{t^\prime} \neq 1$}{
%         $\Delta t = \text{Hop}(h_{t^\prime}, t^\prime)$\;
%         $t^\prime \mathrel{+}= \Delta t$\;
%     }
%     }
%     $\hat{y} = \text{PrefixClassifier}(h_{t^\prime})$\;
%     $\tau = t^\prime$\;
%     \caption{Inference with \method}\label{alg:general}
%     \end{algorithmic}
% \end{algorithm}
% \end{wrapfigure}
% \end{figure}

% \subsection{\method}
We propose an intuitive first solution to the open Early Classification of Irregular Time Series (ECITS) problem, which we name \method and illustrate in Figure \ref{fig:architecture}.
The ultimate goal of our proposed method is to \textit{predict the best halting time} $\tau$ for a given series so as to balance the cost of \textit{delaying a prediction} with the benefits of \textit{accuracy}, according to the requirements of the task at hand.
Thus, one halting timestep $\tau$ is predicted per series $X$ along with a prediction $\hat{y}$ made using only observations made \textit{before} time $\tau$.

Since no ECITS solution exists, we first describe a general solution, which we then instantiate as an architecture that builds on representation learning for irregular time series and on deep reinforcement learning.
A general solution to ECITS iterates three steps: \circled{1} Predict a \textit{candidate} halting time $t^\prime$ given only variables recorded before $t^\prime$. \circled{2} Construct a vector representation $h_t^\prime$ of the ongoing series $X$ that captures patterns in both values and irregularity up to time $t^\prime$. \circled{3} Predict whether or not to halt and classify $X$ at time $t^\prime$.
If so, use $h_{t^\prime}$ to classify $X$. If not, predict the next candidate halting time $t^\prime$.
Thus, a solution will march through the continuous timeline with a step-size that is predicted by the model according to the observations as they arrive.
At each step, the model will decide whether or not to Stop and return a classification.

%---------------------------------------------------------
\begin{table}[t]
  \caption{Basic Notation}
  \vspace{-10pt}
  \centering
  \label{tab:notation}
  \begin{tabular}{ p{1.4 cm} p{6.1 cm}}
    \toprule
    Notation & Description\\
    \midrule
    % $N$                 & Number of time series in dataset.\\
    $D$                 & Variables per time series. \\
    % $C$                 & Number of possible classes.\\
    $\tau$              & Predicted halting time.\\
    $\hat{y}$           & Predicted class label. \\
    $t^\prime$          & Candidate halting time. \\
    $\pi_\text{stop}$   & Stopping policy (chooses Stop or Wait). \\
    $\pi_\text{hop}$    & Hopping policy (chooses hop size). \\
    $h_{t^\prime}$      & Hidden state computed at time $t^\prime$.\\
  \bottomrule
%   &where $i=[1, \dots, k]$, $t=[1, \dots, T]$, and $l=[1, \dots, L]$. \\
%   \bottomrule
\end{tabular}
\vspace{-10pt}
\end{table}
%---------------------------------------------------------

Each component of this general solution solves one challenge of the ECITS problem.
First, learning when to \textit{try} to stop is essential in the irregular setting.
This is in contrast to the standard ECTS setting where, with knowledge that observations arrive on a fixed schedule, methods simply decide whether or not to stop every time a new measurement arrives.
Second, standard supervised learning methods struggle to model ITS data as they are not fixed-length.
Learning dense representations of these data instead provides feature vectors that are easy to learn from.
Step three can then leverage the vast success of deep learning to classify ongoing ITS.

% \wnote{I wonder if we could say a bit more about why this general/modular solutions solves ECISTS, and why this general solution in particular? Like, how each step addresses some challenge, or why each step is there. Step 2) stands out a bit in particular. I'm not sure if the utility of that step is clear as-is (though I get why it's there, just not sure if it's clear enough in the paper currently)}

This modular setup solves the ECITS problem and so we instantiate this idea with solutions to each of the three sub-problems, integrating the two goals of accuracy and earliness. % and following the state-of-the-art by framing this problem as a markov decision process.
% \wnote{slightly unclear - is framing it as a markov decision process part of using state-of-the-art solutions to each step, or you use SOTA and also treat this as a markov decision process? Also, I wonder if we can say something other than expanding on recent SOTA approaches for the sub-problems. Like, subproblem 1) doesn't have a direct analog that's already solved, right? no early classification problem predicts an arbitrary candidate halting time, they just predict either stop or wait (or am I mistaken?)? I wonder if ``expanding" is doing a lot of lifting here, and if we can instead make it seem more like even the solutions to the sub-problems are novel?}
First, a continuous-time recurrent network $R(\cdot)$ constructs a representation $h_{t^\prime} = R(X_{\leq{t^\prime}})$ where $X_{\leq{t^\prime}}$ represents all observations made prior to a timestep $t^\prime$.
Next, a \textit{Halting Policy Network} decides either to \textit{Stop}, or \textit{Hop} forward in time to a new timestep $t^\prime \peq \Delta t$ where $\Delta t$ is a real-valued Hop Size computed as a function of representation $h_{t^\prime}$.
This two-policy setup is novel for the ECTS literature.
Since incoming observations are irregular, our adaptive approach allows the network to learn when to try and stop according to when observations arrive, adding flexibility.
% If the network chooses not to \textit{Stop}, it then also predicts \textit{for how long} to wait before repeating this process, matching the variability in the incoming observations.
During training, the halting policy is encouraged to prefer both \textit{smaller} (earlier) values $\hat{\tau}$ and \textit{accurate} predictions. 
Once the halting policy chooses to stop or the series ends, a classifier network predicts the class label of $X$.

The rest of this section is organized as follows.
We describe our implementation of each component of \method: Generating prefix embeddings with a continuous-time RNN, a neural network for classifying prefix embeddings, and deep reinforcement learning agents for halting early and hopping effectively.
Then, we describe how to train each component. %, culminating with one joint loss function that optimizes the entire system. 

\subsection{Embedding Irregular Time Series Prefixes}\label{sec:GRUD}
\method learns to encode ongoing irregular time series via continuous time representation learning, computing vector representations of a series $X$ at real-valued timesteps.
We refer to this as a \textit{Prefix Encoder}, as it encodes the prefixes of ongoing time series.
% Since most standard supervised learning methods assume fixed-length feature vectors, such representations can facilitate learning from ITS. % \wnote{can we say why continuous time embedding, for people unfamiliar with irregular sampling?}
There has been a recent surge in approaches developed for representing ongoing ITS \cite{che2018recurrent,rubanova2019latent} and most use a recurrent component to encode the series at real-valued timesteps in the continuous timeline:
% There has recently been a wide variety of approaches developed for representing ongoing ITS \cite{che2018recurrent,rubanova2019latent}, which compute encodings $h$ at real-valued timesteps in the continuous timeline:
$h_t^\prime = \mathcal{R}_\theta(X, t^\prime)$, where $R(\cdot)$ is a continuous-time recurrent neural network and $t^\prime$ is a real-valued time. % real-valued point in the continuous timeline of $X$.
$h_t^\prime$ is thus a vector representing all dynamics of observations in series $X$ prior to time $t^\prime$, including information found in the irregularity of the values. % (like density \cite{marlin2012unsupervised}.
The only constraint on architecture design for $\mathcal{R}_\theta(\cdot)$ in the Early Classification setting is that $h_t^\prime$ must only be computed with respect to values observed \textit{earlier} than $t^\prime$.
This disables the use of methods that compute bi-directional hidden states or use future values for imputation \cite{shukla2019interpolation}.
Further, we seek to model the \textit{irregularity itself}, which can inform both the classification accuracy and the earliness.
Thus we compute $h_{t^\prime}$ using the GRU-D \cite{che2018recurrent}, denoted for variable step sizes between embeddings.
The hidden state and input values are decayed based on the time since the
last-observed value for each variable as follows:
\begin{align}
    \hat{x}^d_{t^\prime} &= m_{t^\prime}^dx_{t^\prime}^d \odot (1-m^d_{t^\prime})(\gamma_{x_{t^\prime}}^dx_{\text{prev}}^d + (1-\gamma_{x_{t^\prime}}^d\tilde{x}^d_{t^\prime}))\\
    \hat{h}_{t^\prime} &= \gamma_{h_t} \odot h_{t^\prime}\\
    r_{t^\prime} &= \sigma(W_r\hat{x}_{t^\prime} + U_r\hat{h}_{t^\prime} + b_r)\\
    z_{t^\prime} &= \sigma(W_z\hat{x}_{t^\prime} + U_z\hat{h}_{t^\prime} + b_z)\\
    \tilde{h}_{t^\prime} &= \phi(W\hat{x}_{t^\prime} + U(r_{t^\prime} \odot \hat{h}_{t^\prime}) + Vm_{t^\prime} + b)\\
    h_{t^\prime} &:= (1-z_{t^\prime}) \odot h_{t^\prime} + z_t \odot \tilde{h}_{t^\prime},
\end{align}
where $\tilde{x}^d_{t^\prime}$ is the mean of all values of for the given instance's variable $d$ before time $t^\prime$, $m_{t^\prime}^d$ is a binary value indicating whether or not any new observations have been made since $t^\prime$, and $x_{\text{prev}}^d$ is the value of the most recent observation of the $d$-th variable prior to time $t^\prime$.
$\odot$ is the hadamard product and $\gamma_{x_{t^\prime}}^d$ is a decay factor for variable $d$ at time $t^\prime$ computed by a neural network: $$\gamma_{x_{t^\prime}}^d = e^{-\max(0, W_\gamma s_{t^\prime}^d + b_\gamma)},$$
% \end{equation}
where $s_{t^\prime}^d$ is the difference between $t^\prime$ and the time of the last observation of variable $d$. 
The GRU-D is a natural choice for this problem, as it computes hidden states at any real-valued times, based on previous hidden states.
By incorporating the mask $m$ and time since last observation $x_{\text{prev}}$, the hidden state reflects input irregularity.
When decisions---like when to halt---are made based on the resultant hidden states, they depend on the irregularity of the input series.
Further, the GRU-D is a \textit{unidirectional} state space model, computing hidden states without using future information.
This is crucial, as early classifiers can only use previous observations.

An encoding $h_{t^\prime}$ thus represents knowledge contained in the transitions between the values over time \textit{and} observation density.
These hidden states can therefore use patterns in values and observations to support both earliness and accuracy.
For instance, if observations are arriving rapidly, more may arrive soon, so waiting to receive them will not incur much penalty for waiting \cite{shukla2019interpolation}.
On the other hand, if the hidden states indicate low observation density, it may be better to cut our losses since more observations may be far in the future.
Alternatively, the time-since-last-observation can itself be a valuable feature \cite{lipton2016directly}, so waiting to see if more observations arrive can itself be a reasonable policy for accurate classification.
% Measuring this density is important to both the classification and the earliness goals: Discriminative signals can appear in both the values themselves and their timing \cite{lipton2016directly}.
% The density of measurements can thus indicate the likelihood of observing more relevant information in the near future \cite{shukla2019interpolation}.
% In practice, these update equations can be run multiple times \textit{between} candidate halting times $t^\prime$ to ensure a granular-enough representation of a time series.
% \wnote{should we say why the density is important/that we assume that the density informs the classification or earliness task? or both?}
% To produce hidden state $h_{t^\prime}$ we thus use evenly spaced values $t$.

\subsection{Classifying Prefixes}
Given a prefix embedding $h_{t^\prime}$, we use a \textit{Prefix Classifier} $\mathcal{C}_\theta$ to predict the class label of the entire series $X$ based only on values observed up to time $t^\prime$. %is to classify the entire series $X$ via a \textit{Prefix Classifier} $\text{C}(\cdot)$.
We use a standard fully connected network that projects $h_{t^\prime}$ into a $C$-dimensional probabilistic space via the softmax function, parameterizing the conditional class probabilities.
In our experiments, we use one hidden layer, but this component can be scaled up depending on the task at hand.
Once the Halting Policy Network chooses to stop, the final prediction $\hat{y}$ for series $X$ is also generated by the Prefix Classifier.

\subsection{Irregularity-Aware Halting Policy Network}
\method achieves early halting through an irregularity-aware \textit{Halting Policy Network} $\mathcal{H}_\theta$ that chooses whether to \textit{Stop} or \textit{Wait} at a given time $t^\prime$ based on the history of an ongoing time series $X$.
If it chooses to Wait, it also predicts \textit{for how long} in the form of a real-valued step size, after which it will run again.
Since there are rarely ground truth optimal halting times for when \method \textit{should} stop, we follow the state-of-the-art \cite{hartvigsen2019adaptive,hartvigsen2020recurrent,martinez2019adaptive} and formulate this task as a Partially-Observable Markov Decision Process (POMDP): Given a state $h_{t^\prime}$, select an action $a_{t^\prime}$ from the set \{Stop, Wait\}.
If $a_{t^\prime} = \{\text{Stop}\}$, then the class label for the entire series is returned at time $t^\prime$.
If $a_{t^\prime} = \{\text{Wait}\}$, update $t^\prime := t^\prime + \Delta t$, where $\Delta t$ is a \textit{hop-size} predicted based on $X_{\leq t^\prime}$, the observations before time $t^\prime$.
% Since $X$ is irregular and observations may arrive sporadically, when the Halting Policy Network chooses to \textit{Wait} it must also predict \textit{for how long} before trying to stop again, thereby computing the next time $t^\prime$ at which the next prefix embedding can be computed.
% We refer to this as picking a \textit{Hop Option}.
Overall, the earlier and more accurate the predictions are for a series $X$, the higher the reward.
If \method stops too early, it may not have observed enough data and will be less likely to be accurate.
Therefore, a good solution must carefully balance between stopping and predicting the next halting time.

\textbf{States.}
The Halting Policy Network's first job is to compute a probability of halting $p_{t^\prime}$ given the environment's state at time $t^\prime$ which can then be used to take an action: Stop now, or Wait for more data to be collected.
While prior methods use only the prefix embedding $h_{t^\prime}$ to represent the ongoing system's current state \cite{hartvigsen2019adaptive}, we propose also predicting an intermediate classification $y_{t^\prime}$ via the \textit{Prefix Classifier} for each hidden state to gauge the model's current opinion of the classes. % prior to predicting whether or not to Stop.
By feeding this information to the Halting Policy Network, it can learn the relationship between the Prefix Classifier's confidence and the likely accuracy, which often depends on the task.
Additionally, we input $t^\prime$ as additional knowledge of a prediction's \textit{earliness}.

\textbf{Actions.}
The Halting Policy Network contains two prediction problems in sequence. First, it decides whether to \textit{Stop} and classify, or to \textit{Hop} forward in time and wait to classify.
For the \textit{Stop} decision, the model selects actions from the set $\mathcal{A}_\text{stop} = \{\text{Stop}, \text{Hop}\}$.
% While values $\delta$ may be hand-picked if domain knowledge is available, we simplify by picking a value $\delta_0$, then computing further values sequentially: $\delta_i = \delta_{i-1} + \delta_0$.
% This sacrifices some fine-grained control over the action space in favor of practicality.
% In principle, this assumption may be relaxed either by considering more hop options or sampling from a continuous distribution.
We thus parameterize a binomial distribution over the action set $\mathcal{A}_\text{stop}$ by a stopping policy $\pi_\text{stop}$:
% Using the softmax function, probability $p_t$ then parameterizes a multinoulli distribution over $H+1$ categories where $H$ is the number of \textit{hop options}:
\begin{align}
    p_{t^\prime} = \text{softmax}(W_h h_{t^\prime} + U_h \hat{y}_{t^\prime} + V_h t^\prime + b_h),
\end{align}
where $W$, $U$, and $V$ are weight matrices and $b$ is a bias vector.
Finally, we use the probabilities $p_{t^\prime}$ to sample an action from a multinomial distribution. % from which we sample action $a_{t^\prime}$.
If $a_{t^\prime} = \{\text{Stop}\}$, then the corresponding class prediction $\hat{y}_{t^\prime}$ is returned at time $t^\prime$ and we set $\tau$ is set to $t^\prime$.

If the model instead chooses to \textit{Hop}, we run a hopping policy $\pi_\text{hop}$, another small neural network, that predicts a positive real-valued hop-size, which is added to $t^\prime$.
To account for the irregular nature of the input time series, we model the hop-time as a continuous variable, samples of which we acquire by parameterizing a normal distribution with a neural network.
% Since we are working with the continuous timeline, this hop-time is naturally modeled using continuous values, which we sample from a parameterized normal distribution.
The hop policy $\pi_\text{hop}$ begins by predicting the a mean value:
\begin{align}
    \mu_{t^\prime} = \phi(W h_{t^\prime} + U \hat{y}_{t^\prime} + V t^\prime + b),
\end{align}
where $\phi$ is the ReLU function \cite{hahnloser2000digital}.
% $\mu = \sigma(Wh_{t^\prime} + b)$, 
We then sample a hop-size $\Delta t$ from the normal distribution with mean $\mu_{t^\prime}$: $\Delta t \sim \mathcal{N}(\mu_{t^\prime}, \sigma)$.
We leave the standard deviation $\sigma$ as a hyperparameter, though in principle it can also be learned by the model.
To ensure $\Delta t \geq 0$, we take the absolute value of $\Delta t$, which is a common approach in similar scenarios \cite{mnih2014recurrent}. 
% In our experiments, we set the standard deviation to $0.1$, indicating 10\% of the timeline.
To compute the new candidate halting time, we add the hop-size to the current candidate halting time: $t^\prime \mathrel{{+}{=}} \Delta t$.

In practice, early in training the \textit{Halting Policy Network} may tend to exploit some actions by predicting too-high probabilities, relative to an optimal policy.
To encourage exploration early on in training, we employ a simple $\epsilon$-greedy approach to action selection and exponentially decay the values $\epsilon$ from 1 to 0 throughout training:
\begin{align*}\label{eqn:epsilon_greedy}
    a_{t^\prime} =
    \begin{cases}
        a_{t^\prime}, &\text{with probability}\ 1-\epsilon\\
        \text{random action,} & \text{with probability}\ \epsilon
    \end{cases}
\end{align*}
By exploring more, the model tries out different sequences of actions to cover the space of possible episodes more effectively while early on in training.
We also increase the probability that the Halting Policy Network chooses to \textit{Wait} before the model has been thoroughly trained so that the prefix embeddings and classifier also get to observe more of the sequences and increase their performance.
Otherwise, a model that learns to stop early very quickly will never have seen the later portions of the training sequences.
For all of our experiments, we compute $\epsilon$ as $e^{-i}$ while reassigning  $i := i + \frac{i*7}{E}$ for $E$ training epochs after initializing $i$ to 0.
During testing, we set $\epsilon=0$ to avoid exploration for testing series. %so that there is no random exploration for testing series.

\textbf{Rewards.}
The final component of the POMDP is the \textit{reward} for reaching different states.
We encourage the halting policy network to cooperate with the prefix embeddings and classifier by setting the reward $r_{t^\prime}=1$ when the $\hat{y}$ is accurate and setting $r_{t^\prime}=-1$ otherwise.
This way, accurate classifications are encouraged.
To encourage early classifications, we penalize Wait probabilities, as discussed in the next section, which describes how we jointly train all of \method's components.
% During optimization, we use the standardized and discounted sum of future rewards with a discount factor of $0.99$. 
% \wnote{i might be wrong, but should we also say that the reward depends on earliness?}

\subsection{Training}
The prefix embedding network $\mathcal{R}_\theta$ and Classifier $\mathcal{C}_\theta$ are fully-differentiable, so we train them together using standard back propagation.
We encourage them to predict $\hat{y}$ as close to $y$ as possible by minimizing their cross entropy, where $y_c=1$ if $y=c$ and is 0 otherwise in Equation \ref{eqn:CE}. $\hat{y}_c$ is the predicted probability of class $c$. % indicates when $y$ is class $c$:
\begin{equation}\label{eqn:CE}
    % \mathcal{L}_\text{acc} = \sum_{c=1}^C-y^c\log(\hat{y}^c) + (1-y^c)\log(1-\hat{y}^c)
    \mathcal{L}_\text{acc} = -\sum_{c=1}^C y_c\log(\hat{y}_c)
\end{equation}

The Halting Policy Network, on the other hand, samples its actions and so its training is more intensive, though we follow the standard policy gradient method and use the REINFORCE algorithm \cite{williams1992simple} to estimate the gradient with which we update the network's parameters.
To balance between earliness and accuracy, the parameters of the Halting Policy Network are updated with respect to two goals: Make $\tau$ small, and make $\hat{y}$ accurate.
Following the state-of-the-art \cite{hartvigsen2019adaptive}, we achieve smooth optimization by rewarding accurate predictions and penalizing \textit{the cumulative probability of Waiting} over each episode.
Thus the loss function for optimizing the halting policy network is computed as:
% \begin{align*}
%     \mathcal{L}_\text{hpn} =
%     &-\mathbb{E}\left[\sum_{t^\prime}\log\pi(a_{t^\prime}|h_{t^\prime})\left[\sum_{j=t^\prime}(r_{t^\prime}-b_{t^\prime})\right]\right]\\
%     &- \lambda\sum_{t^\prime}\log\pi(a_{t^\prime}=\text{Stop}|h_{t^\prime}),
% \end{align*}
\begin{align*}
    \mathcal{L}_\text{early} =
    &-\mathbb{E}\left[\sum_{t^\prime}\log\pi_\text{stop}(a_{t^\prime}|h_{t^\prime})\left[\sum_{j=t^\prime}(r_{t^\prime}-b_{t^\prime})\right]\right]\\
    &-\mathbb{E}\left[\sum_{t^\prime}\log\pi_\text{hop}(a_{t^\prime}|h_{t^\prime})\left[\sum_{j=t^\prime}(r_{t^\prime}-b_{t^\prime})\right]\right]\\
    &- \lambda\sum_{t^\prime}\log\pi_\text{stop}(a_{t^\prime}=\text{Stop}|h_{t^\prime}),
\end{align*}
where the scale of $\lambda$ determines the emphasis on earliness and $\pi_\text{stop}$ and $\pi_\text{hop}$ are the probabilities predicted by the stopping and hopping networks, respectively.
If $\lambda$ is large, the halting policy network will learn to maximize the probability of stopping always whereas if $\lambda$ is small or zero, the model will solely maximize accuracy.
Interestingly, the most-accurate classification may not always be achieved by observing the entire series, though this is rare in practice.
For example, early signals followed by irrelevant values make classification challenging for memory-based models.
Our approach deals with this case naturally by learning not to make late predictions when they are less accurate, regardless of any cost of delaying predictions.
% \wnote{should we point out that in this (rare?) case the goals of earliness and accuracy are to an extent aligned? though this goes against one of the challenges}
The final loss $\mathcal{L}$ is thus 
\begin{equation}\label{eqn:alpha}
\mathcal{L} = \mathcal{L}_\text{acc} + \alpha\mathcal{L}_\text{early},
\end{equation}
and minimized via gradient descent. $\alpha$ scales the loss components.

% \wnote{I'm stopping here for tonight; i'll continue with the Experiments section tomorrow}

% This modular framework solves the ECISTS problem and in this work we thus instantiate this idea by building on recent state-of-the-art approaches to each of the three sub-problems.
% First, we use a continuous-time RNN $R()$ to estimate $h_{\hat{\tau}}$ at any real-valued timestep (\textit{i.e.}, $h_{\hat{\tau}}$ = $R(X_{\leq\hat{\tau}})$). Second, we use a \textit{Halting Policy Network} that observes the hidden state $h_{\hat{\tau}}$ and decides whether or not to classify $X$ at time $\hat{\tau}$. During training, the halting policy network is encouraged to prefer \textit{smaller} (earlier) values $\hat{\tau}$.
% This network is modeled as a reinforcement learning agent that discretely selects whether or not to halt.
% If it chooses \textit{not} to halt, it then predicts how long to wait before repeating this process.

% : Construct a vector representation $h$ for an ongoing series $X$ up to a \textit{candidate} halting time $\hat{\tau}$. Then, decide whether or not this representation effectively discriminates between the given set of classes. If so, stop and predict. If not, Wait for more observations. Since ISTS data naturally arrive at irregular times, we also learn \textit{for how long} to Wait before trying to stop again.

\section{Experiments}
\label{sec:experiments}
% We evaluate \method using both synthetic and real-world data, ultimately demonstrating that our approach succeeds to identify the appropriate halting times for previously-unseen time series.

% \subsection{Compared Methods}
% \begin{itemize}
%     \item \textbf{Preset Halting.}
%     \item \textbf{Classification Confidence Thresholding.}
% \end{itemize}

\subsection{Datasets}\label{sec:data}

We evaluate \method on four synthetic and three real-world datasets, which are described as follows:
% \wnote{it might be better to separate out the four synthetic datasets more. right now, we say there are four synthetic datasets and 3 real-world datasets. Then we list four datasets (saying 4 based on the textsc headers): 1 synthetic (that is actually a group a 4 datasets), and 3 real. Not 100\% sure this should change (listing out the 4 synthetic in separate paragraphs might not be worth it, since it'd take up a lot of space and maybe they're somewhat similar. But maybe wrap ``Uniform", ``Early-Normal", etc in textsc even if they're within 1 paragraph?}

% \wnote{also, I don't see ``Simple Signal" in any of the figures. If i'm quickly skimming through here, it looks like there's no results on this dataset (even though I know figure 2 shows the 4 datasets represented by SimpleSignal, of course)}

% \jida{I agree that we need the other way to show the 4+3 datasets. Right now the 4 is too hidden and also the 2 for extrasensory. How about: \\
% \textbf{Four Synthetic Datasets}: 
% Then each of them in the paragraph are emphasized by textsc. So that we mean all textsc fonts refer to one dataset.
% \\
% \textbf{Three Real-World Datasets}:
% Then separate into 3 subparagraphs in corresponding to Figure 4. \\
% \textsc{ExtraSensory Walking}:\\
% \textsc{ExtraSensory Running}:\\
% \textsc{PhysioNet}:
% }

\textit{Synthetic datasets}:
Since the true halting times are not available for real data, we develop four synthetic datasets with known halting times.
% Each has its true signal signal times sampled from a preset distribution. % a different distribution from which true signal timings are sampled.
Intuitively, a good early classifier will stop as soon as a signal is observed.
To generate these data, we first uniformly sample $T-1$ timesteps from the range $[0, 1]$ for each of $N$ time series.
We set $T=10$ and $N=5000$.
Then, from a chosen distribution of signal times, we sample one more timestep per series at which the class signal occurs and add it to a series' timesteps. 

We experiment with four distributions of true signal times, each creating a unique dataset: \textsc{Uniform} $\mathcal{U}(0, 1)$, \textsc{Early} $\mathcal{N}(\mu=0.25, \sigma=0.1)$, \textsc{Late} $\mathcal{N}(\mu=0.75, \sigma=0.1)$, and \textsc{BiModal} where half the signals are from \textsc{Early} and the other half are from \textsc{Late}.
For each time series, one value is sampled from one of the distributions---each distribution creates one dataset---which serves as the time at which a signal arrives in the timeline.
In all cases, we clamp this value to be within range $[0, 1]$.
We generate two classes by giving $2500$ time series a $1$ at their signal occurrence time, and $-1$ to the remaining $2500$ series in the dataset. % have half of the values at the signal occurrence times to be $1$ and the other half to be $-1$.
% \jida{This is not clear to me too. Do you mean each class has half of 1 and half of -1? Then how we know the difference between the two classes. Is this ``class" the same as label later that is assigned one per time series?}
%that corresponds to the true signal locations to be $1$ and the other half to be $-1$.
thus we know precisely when the signals arrive for each instance.
Values for off-signal timesteps are set to 0 and are uniformly sampled from the timeline.
% One label is associated with each time series. % and the signal indicating the class appears at one timestep.
% \jida{I am obviously confused. I think because I cannot follow well that I don't know if the description is talking about the dataset, the time series or the signals in each series.}

\textsc{ExtraSensory}: %\footnote{http://extrasensory.ucsd.edu/}
We use the publicly-available ExtraSensory \cite{vaizman2017recognizing} human activity recognition dataset, which contains smartphone sensor data collected across one week while participants labeled which actions they performed and when.
% Participants downloaded mobile applications that logged their movement (e.g., accelerometer or gyroscope), then recorded what activities they did when.
These data were collected in the wild, so participants were left to their own devices for the duration of the study with no prescribed behavior.
Using these data, we simulate a listening probe on a smartphone's accelerometer data, which consist of three variables corresponding to the X, Y, and Z coordinates.
A listening probe saves a phone's battery by collecting data only when certain measurements are taken, naturally creating irregular time series.
For this dataset, we measure the norm of the 3-dimensional accelerometer data, only taking measurements associated with changes in the norm over 0.001.
Here we consider the popular tasks of detecting \textsc{Walking} and \textsc{Running}, classifying whether or not a person performs an activity within a window.
Since activity records are often incomparable between people, we use the records from the person who walked or ran the most breaking their series into 100-minute long windows.
This creates two independent datasets, one per user, which is common for human activity recognition.
We balance each dataset, resulting in 2636 time series with an average of 90 observation per series for \textsc{Walking} and 3000 time series with on average 100 observations for \textsc{Running}.
Deeper, extended discussion of this dataset is available in a concurrent submission.
In general, down-sampling human activity recognition data is common in the irregular time series literature \cite{shukla2019interpolation,rubanova2019latent}, and simulating a listening probe adds a key new ingredient: non-random down-sampling.
%We finally balance the data, resulting in 2636 irregularly-sampled time series with an average of 99 observations per series.

\textsc{PhysioNet}:
The \textsc{PhysioNet} dataset \cite{silva2012predicting} contains medical records collecting from the first 48 hours after 4000 patients were admitted to an intensive care unit and is publicly-available. %\footnote{https://physionet.org/content/challenge-2012/1.0.0/}
% www.physionet.org/content/challenge-2012/view-license/1.0.0/}.
There are 42 variables recorded at irregular times for each patient along with one label indicating if they perished.
This is a common benchmark for multivariate irregular time series classification \cite{che2018recurrent}.
On these data, we train our classifiers to perform mortality prediction for previously-unseen patients.
13.8\% of patients have positive labels, so we use the Area Under the Receiver-Operator Curve (AUC) as our primary metric on all three real-world datasets.

\subsection{Compared Methods}
We compare \method to the two key alternatives.
Each is a state-of-the-art early classifier which we update to handle ITS.
\begin{itemize}
    \item \textit{E-GRU} \cite{dennis2018multiple}. E-GRU thresholds a sequential classifier's output probability in real time. When the predicted probability $\hat{y}$ surpasses a threshold $\alpha$, $\hat{y}$ is used to classify the series, ignoring all future observations. A hidden state represents the streaming series and is updated whenever a new observation arrives. Each time it is updated, the hidden state is passed to a neural network that predicts $\hat{y}$.
    \item \textit{EARLIEST} \cite{hartvigsen2019adaptive}. Similar to E-GRU, EARLIEST models ongoing time series with an RNN. Whenever a new observation arrives, a new hidden state is computed. Similar to \method, a halting policy then decides whether or not to stop and classify the series with a neural network. This baseline also helps to ablate the Hopping Policy Network network, as it only chooses between Stop and Wait, ignoring irregularity between observations.
\end{itemize}

In our synthetic experiments, we also compare \method with a baseline: A GRU is trained to classify irregular time series, then forced to stop at different proportions of the timeline.
This is the most basic early classifier as the halting times are always the same, disregarding the input data.
For all methods, we use the GRU-D update equations \cite{che2018recurrent} described in Section \ref{sec:GRUD}. 

\begin{figure}[t]
    \centering
    \begin{subfigure}{0.98\linewidth}
        \includegraphics[width=\textwidth]{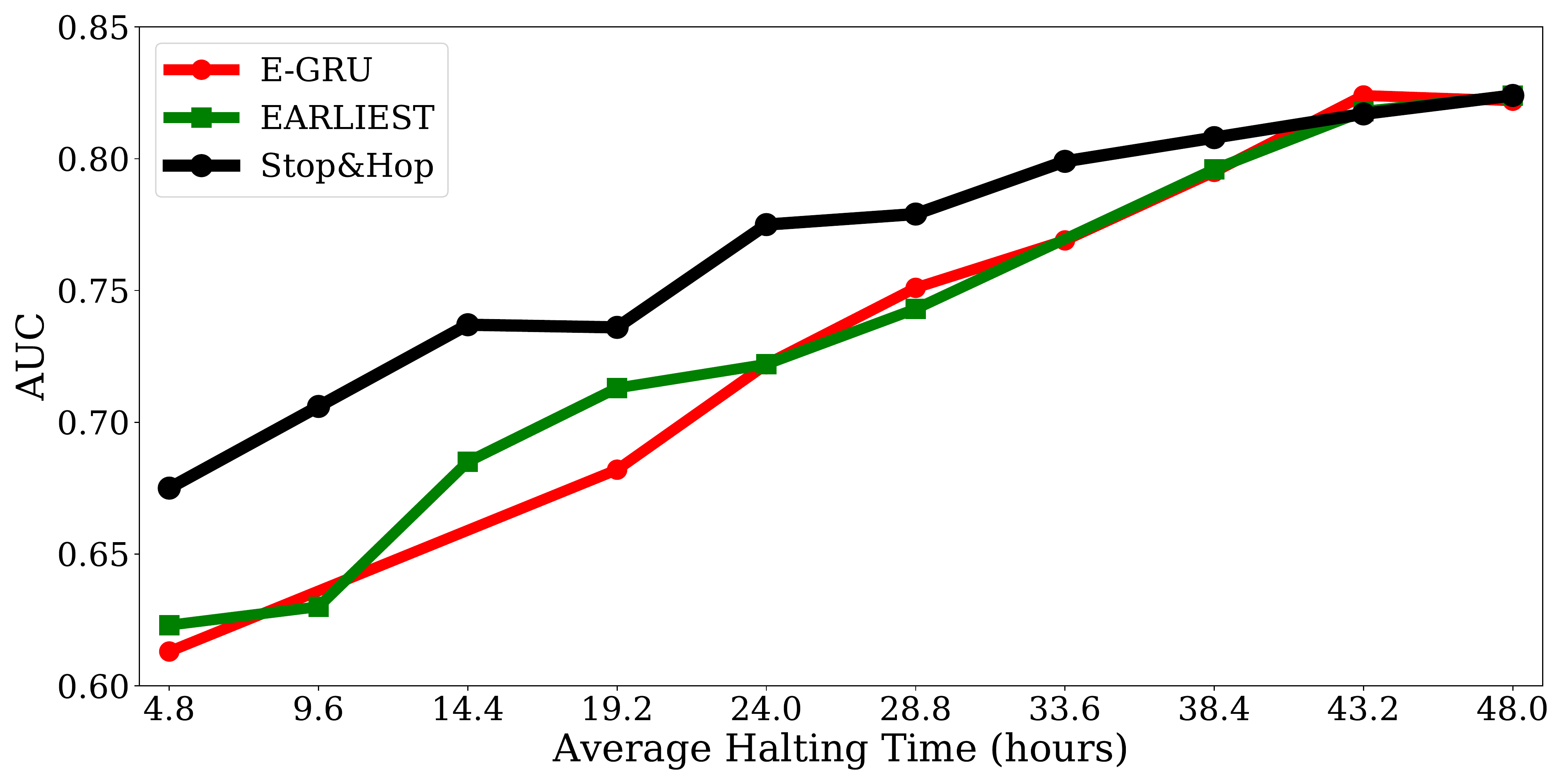}
        \vspace{-5mm}
        \caption{\textsc{PhysioNet}}
    \end{subfigure}
    \begin{subfigure}{0.98\linewidth}
        \includegraphics[width=\textwidth]{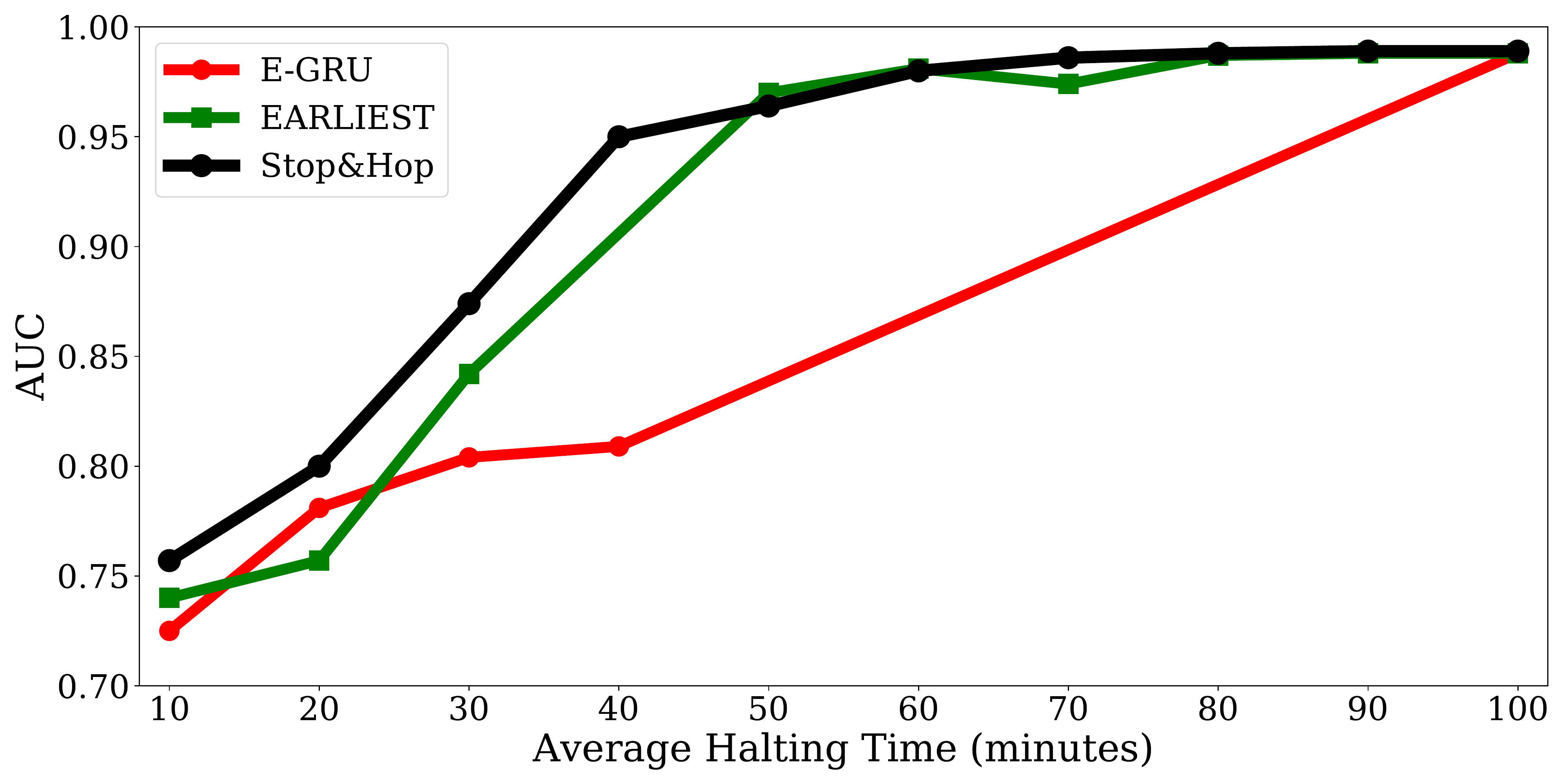}
        \vspace{-5mm}
        \caption{\textsc{ExtraSensory Running}}
    \end{subfigure}
    \vspace{2mm}
    \begin{subfigure}{0.98\linewidth}
        \includegraphics[width=\textwidth]{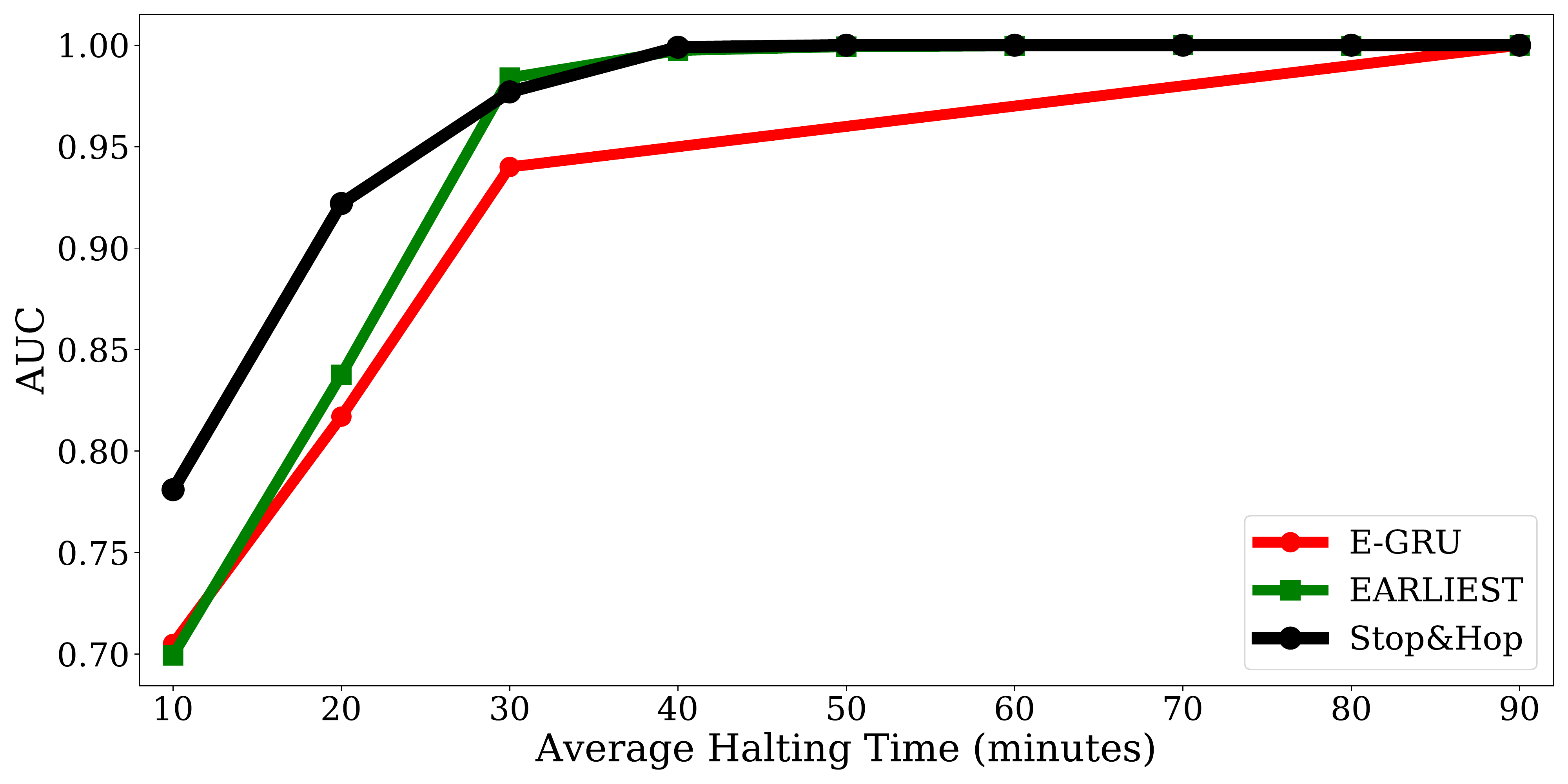}
        \vspace{-5mm}
        \caption{\textsc{ExtraSensory Walking}}
    \end{subfigure}
    \vspace{2mm}

    \vspace{-3mm}
    \caption{Trade-off between earliness and accuracy on three real-world time-sensitive datasets.
    For each dataset in (a)-(c), the black line (\method) outperforms the compared methods, as it is above them for most halting times.
    }
    \label{fig:real_results}
    \vspace{-5mm}
\end{figure}

\subsection{Implementation Details}
For our synthetic datasets and \textsc{PhysioNet}, we repeatedly split the data into 90\% training and 10\% testing subsets five times, and report the average performance. %and hold out 90\% training and 10\% testing splits,standard 80\% training, 10\% validation, and 10\% testing split.
We learn each model's parameters on the training set, and report all final evaluation metrics on the testing set.
The \textsc{ExtraSensory} datasets, on the other hand, contain instances taken from different windows along a single timeline and so we select a timestep for each before which is the training/validation data and after which is the testing to ensure the testing set's sanctity. % and so data splitting is performed \textit{in the timeline} by selecting some timestep before which we extract windows of training data and after which we collect windows of testing data.
% Purely random window-selection would lead to severe cross-contamination between training and testing sets.
% To circumvent this issue, we perform its training and testing splitting \textit{in time}.
% Remaining details are provided in Appendix \ref{sec:har}.
% The training/testing process is repeated five times and we report the average and standard deviation for all experiments.
For the synthetic \textsc{SimpleSignal} datasets, we use 10-dimensional prefix embeddings, which we compute at intervals of 0.1.
% For the \textsc{ExtraSensory} and \textsc{PhysioNet} datasets, we use 20-dimensional representations.
For \textsc{ExtraSensory}, we use 50-dimensional prefix embeddings and for \textsc{PhysioNet} they are 20-dimensional.

\begin{figure}[t]
    \centering
    \begin{subfigure}{.98\linewidth}
        \includegraphics[width=\textwidth]{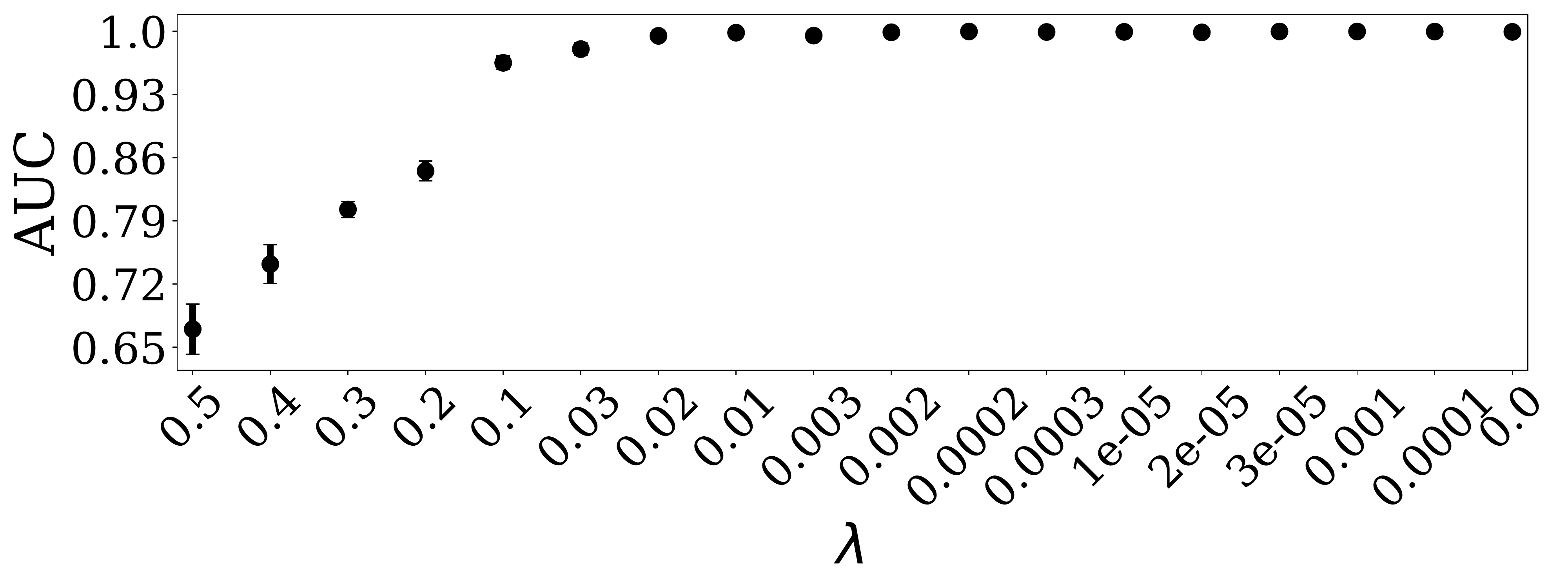}
        \vspace{-5mm}
        \caption{Effect of $\lambda$ on AUC}
    \end{subfigure}
    \begin{subfigure}{.94\linewidth}
        \includegraphics[width=\textwidth]{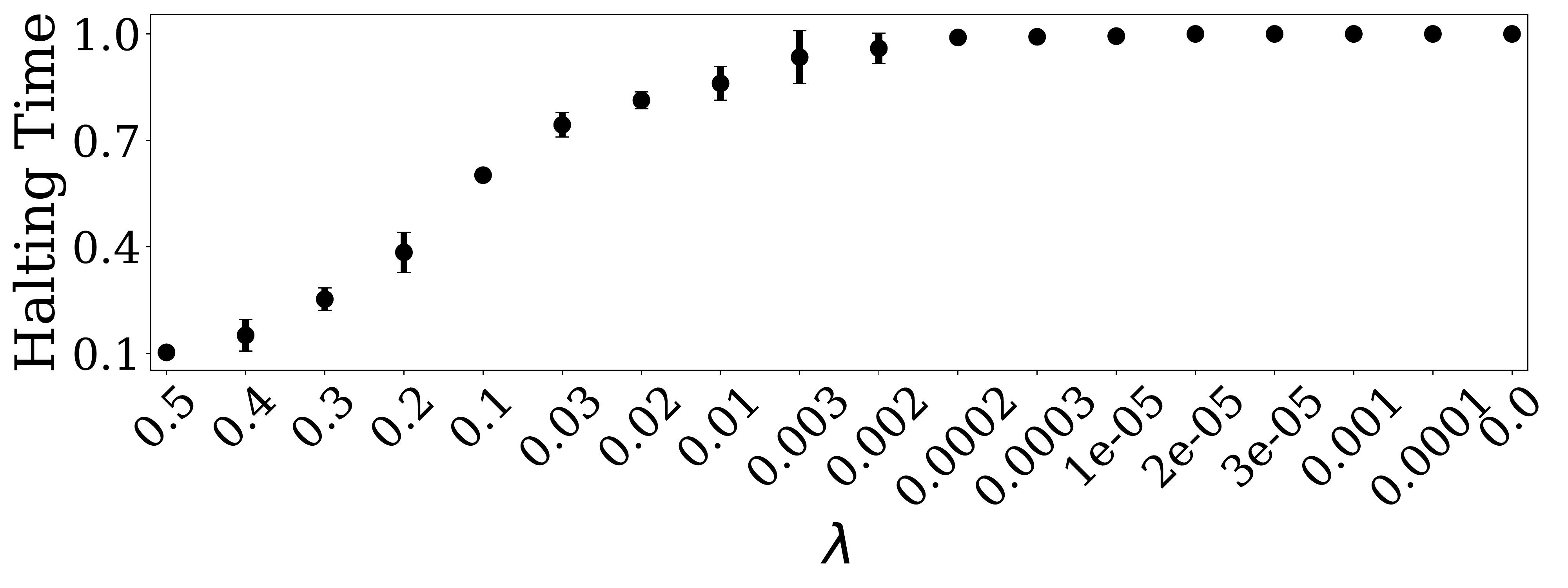}
        \vspace{-5mm}
        \caption{Effect of $\lambda$ on Earliness}
    \end{subfigure}
    \caption{Hyperparameter tuning for $\lambda$, the emphasis on earliness, on the \textsc{ExtraSensory Running} dataset. In (a), the AUC increases smoothly as $\lambda$ decreases. Similarly, in (b) as $\lambda$ decreases, the resultant halting time increases, successfully covering the whole space from 0.1 to 1.0.}
    \label{fig:hyperparams}
    % \vspace{-5mm}
\end{figure}

\begin{figure*}[t]
    \centering
    \begin{subfigure}{0.32\linewidth}
        \includegraphics[width=\textwidth]{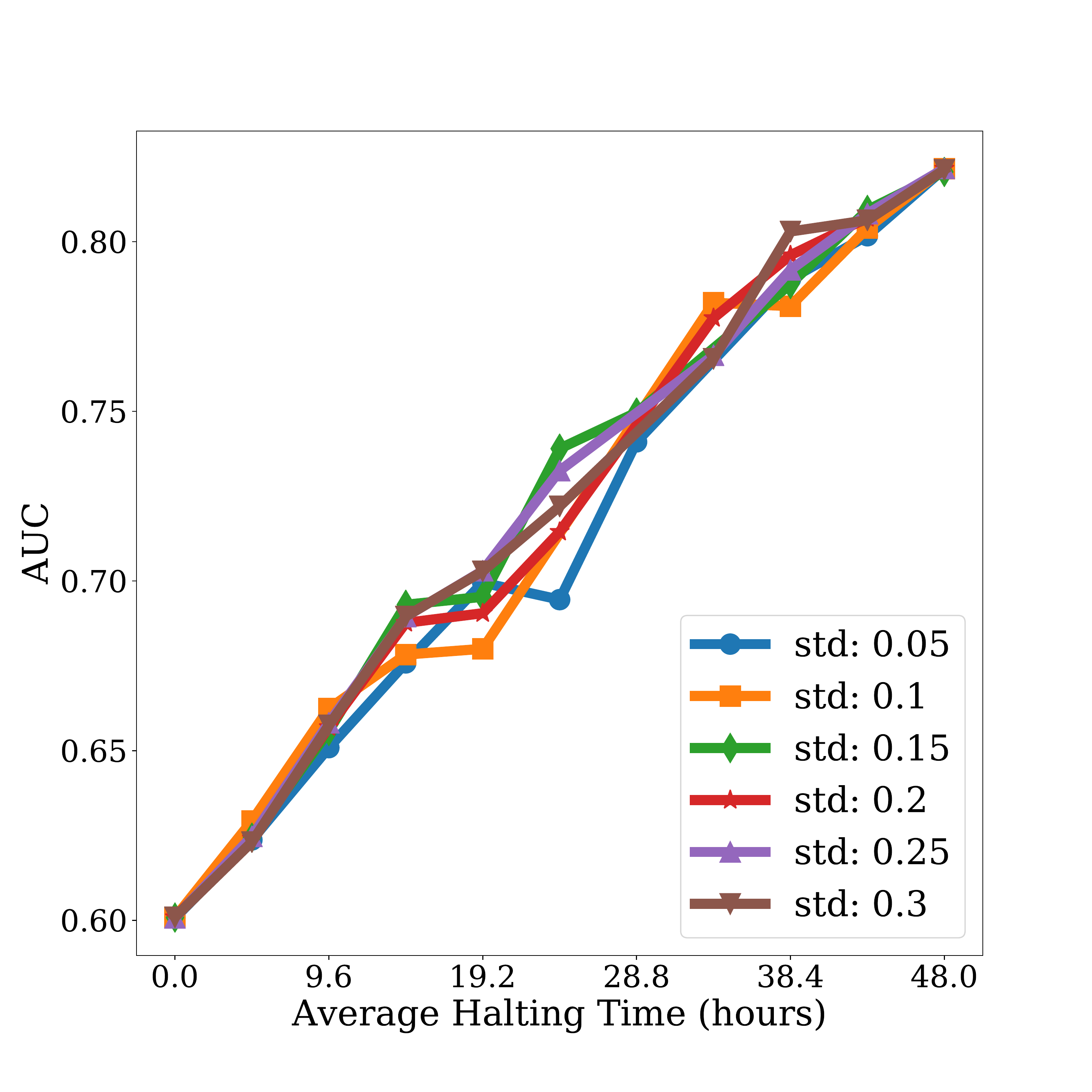}
        \vspace{-5mm}
        \caption{\textsc{PhysioNet}}
    \end{subfigure}
    \begin{subfigure}{0.32\linewidth}
        \includegraphics[width=\textwidth]{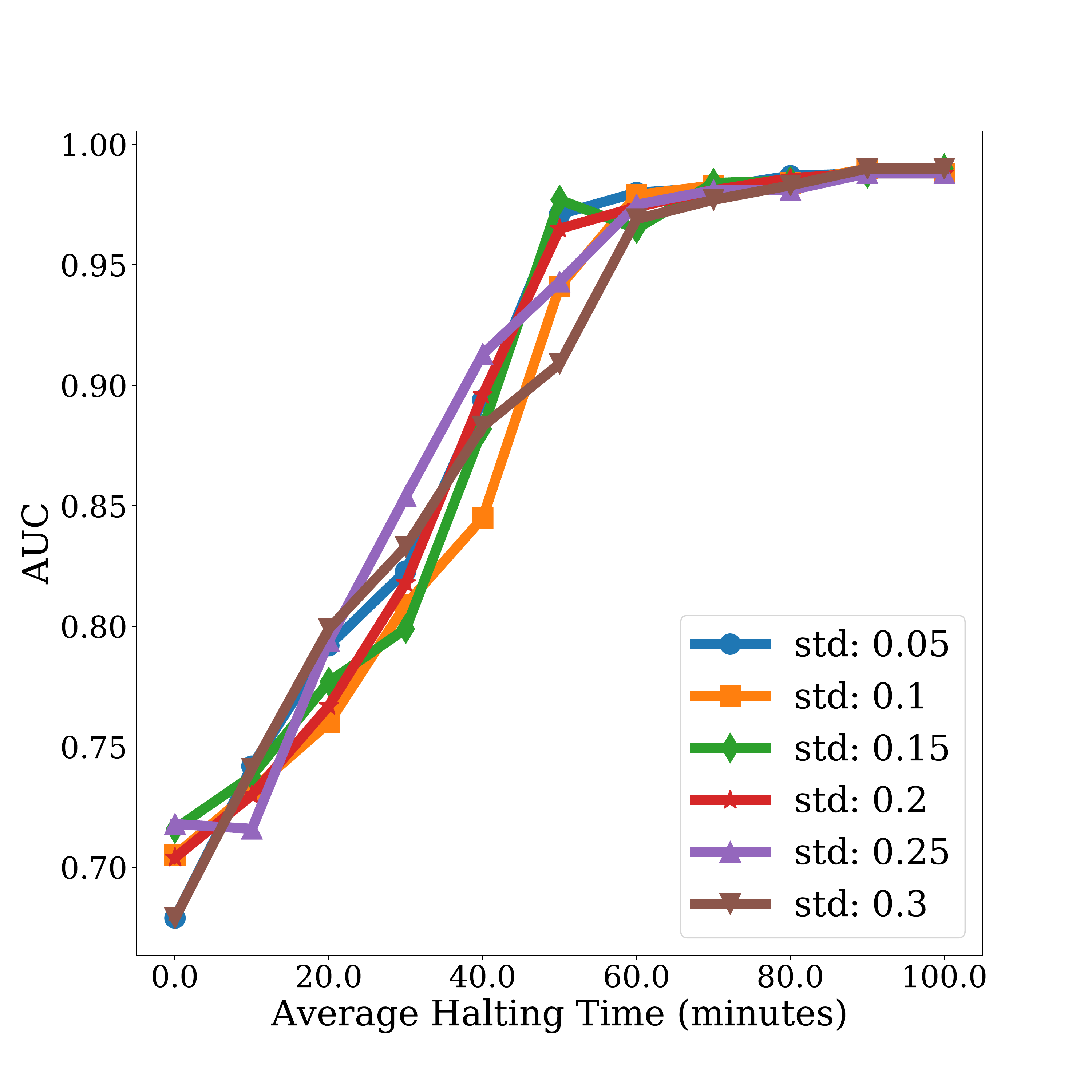}
        \vspace{-5mm}
        \caption{\textsc{ExtraSensory Running}}
    \end{subfigure}
    \vspace{2mm}
    \begin{subfigure}{0.32\linewidth}
        \includegraphics[width=\textwidth]{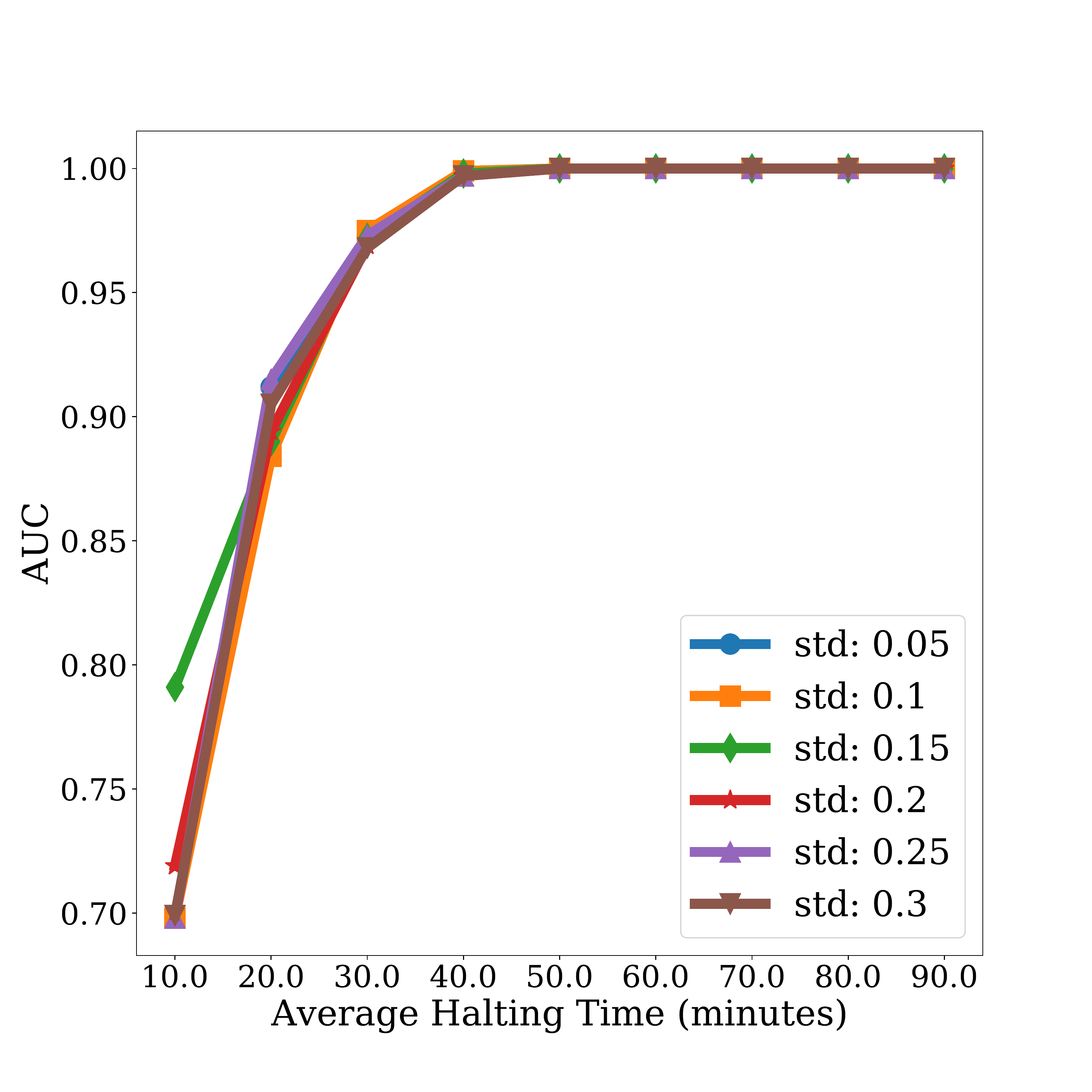}
        \vspace{-5mm}
        \caption{\textsc{ExtraSensory Walking}}
    \end{subfigure}
    \vspace{2mm}

    \vspace{-3mm}
    \caption{Effect of Hopping Policy's standard deviation on the earliness--accuracy trade-off curves. As the curves are similar to one another, \method is generally robust to selecting the standard deviation, though tuning per dataset can still be valuable.
    }
    \label{fig:std_hyper}
\end{figure*}

For each method, we use a batch size of 32 and grid search for a learning rate (options: $\{1e^{-2}, 1e^{-3}, 1e^{-4}\}$) and weight decay for L2 regularization (options: $\{1e^{-3}, 1e^{-4}, 1e^{-5}\}$) using our validation data.
The validation data is a random 10\% of the training dataset and we repeat this random splitting five times.
In our experiments, since we use a GRU-D \cite{che2018recurrent} to compute prefix embeddings, its hidden state should be updated in between hop sizes. For simplicity, we only update the embeddings when real data are observed per the baselines, though \method's final Stop time may be between observations.
Each model was optimized using Adam \cite{kingma2014adam} with learning rates and weight decays that maximize their performance on the validation data.
All training was done using Intel Xeon Gold 6148 CPUs.
For all experiments, we set the value $\alpha$ in Equation \ref{eqn:alpha} to 1 as the loss terms are balanced.
All of our code and data are publicly available.\footnote{\url{https://github.com/thartvigsen/StopAndHop}}

\subsection{Results on real-world datasets}
First, we demonstrate that \method produces early and accurate classifications using three real time-sensitive datasets, shown in Figure \ref{fig:real_results}.
We use the trade-off curves between earliness and AUC to measure each method's performance, adjusting their parameters to achieve average halting times that span the timeline.
% For \method, this means searching for $\lambda$ values.
We first observe that \method outperforms the comparisons: the black line is consistently highest.
This improvement indicates that \method stops earliest and makes the most-accurate classifications.
% By improving over EARLIEST, we can be sure the Hopping policy is integral to \method's strong performance.

\begin{figure*}[t]
    \centering
    \begin{subfigure}{.32\linewidth}
        \includegraphics[width=\textwidth]{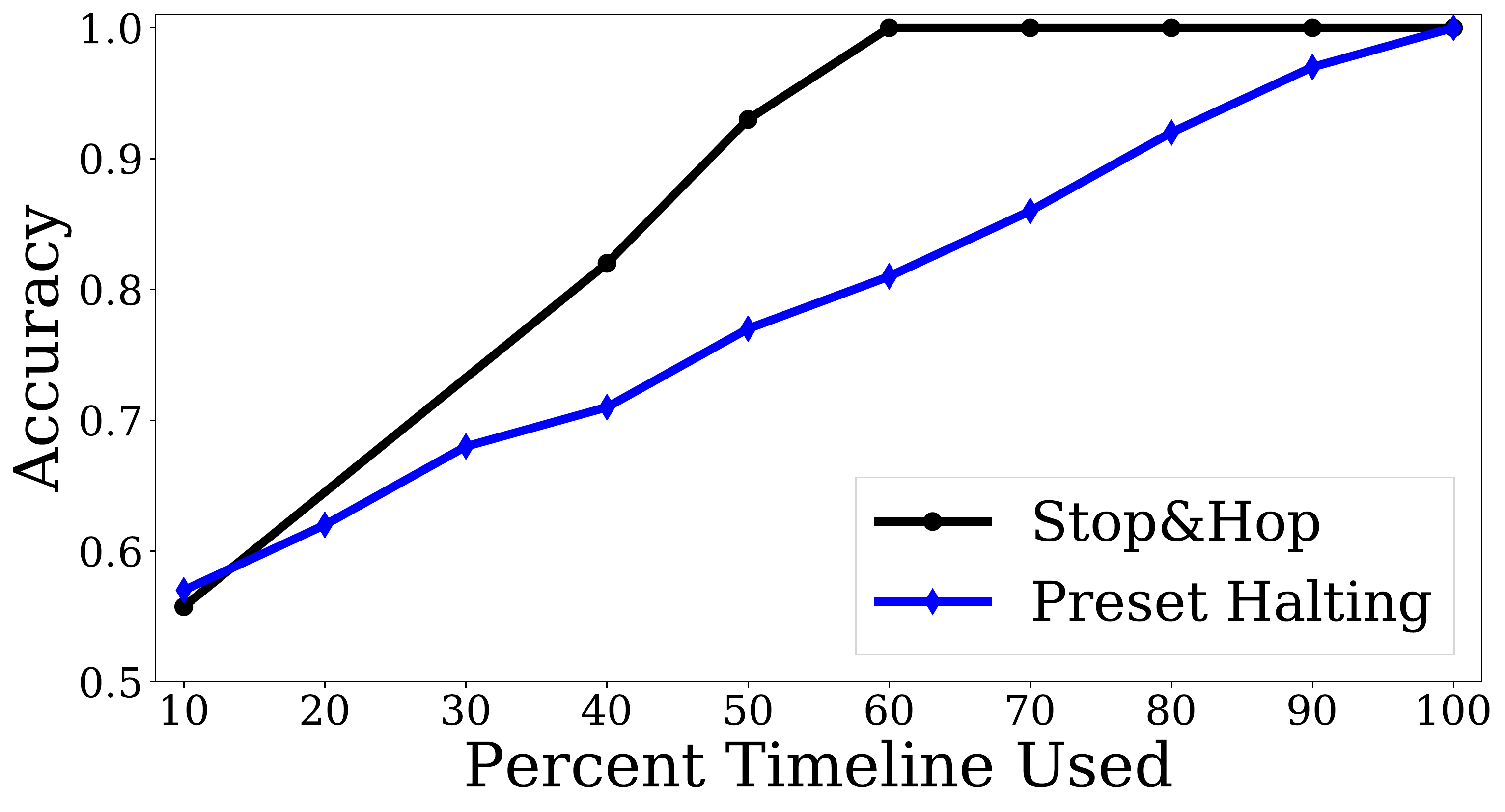}
        \vspace{-5mm}
        \caption{\textsc{Uniform} true signal times}
    \end{subfigure}
    \begin{subfigure}{.32\linewidth}
        \includegraphics[width=\textwidth]{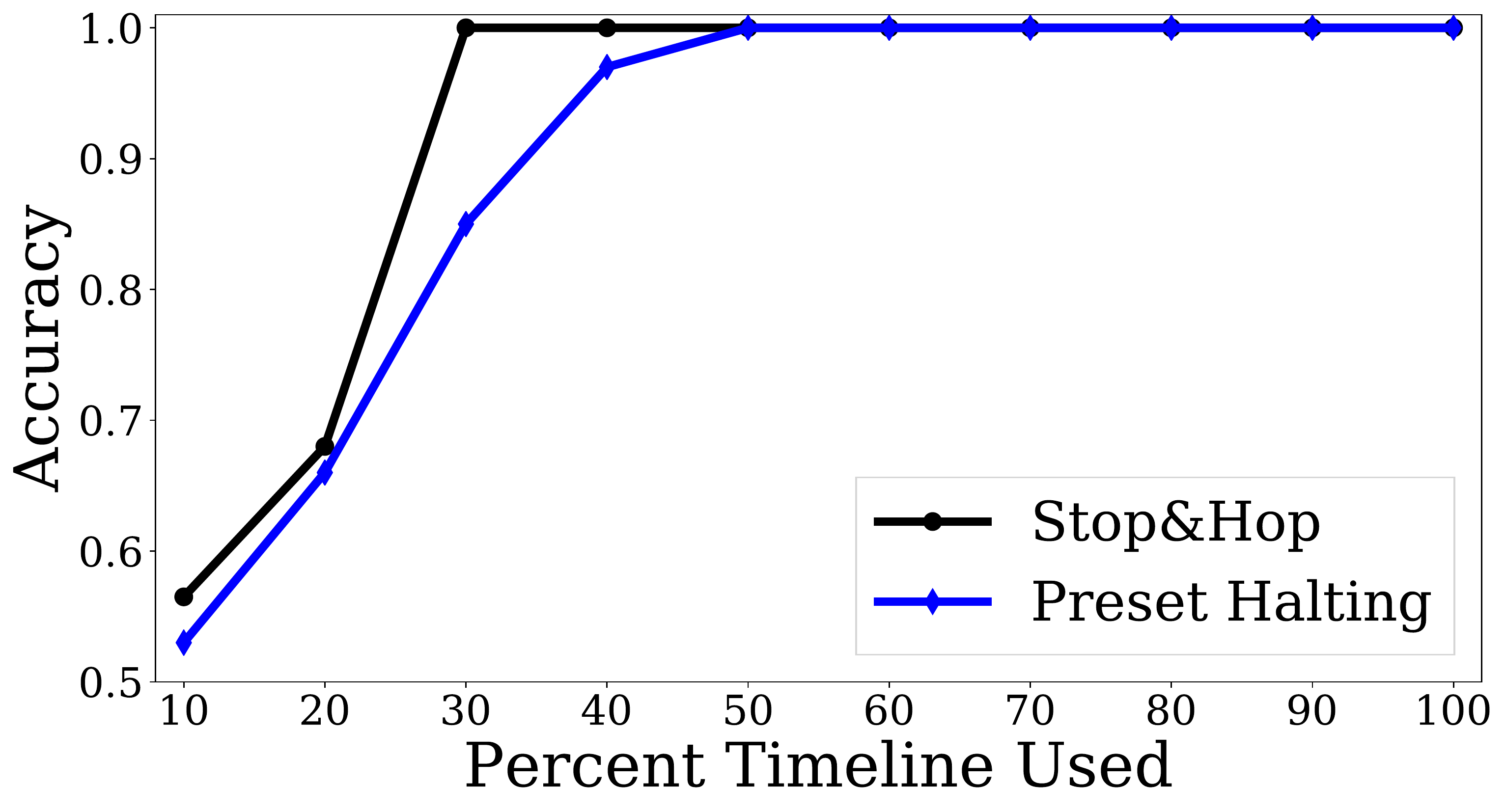}
        \vspace{-5mm}
        \caption{\textsc{Early} true signal times}
    \end{subfigure}
    \begin{subfigure}{.32\linewidth}
        \includegraphics[width=\textwidth]{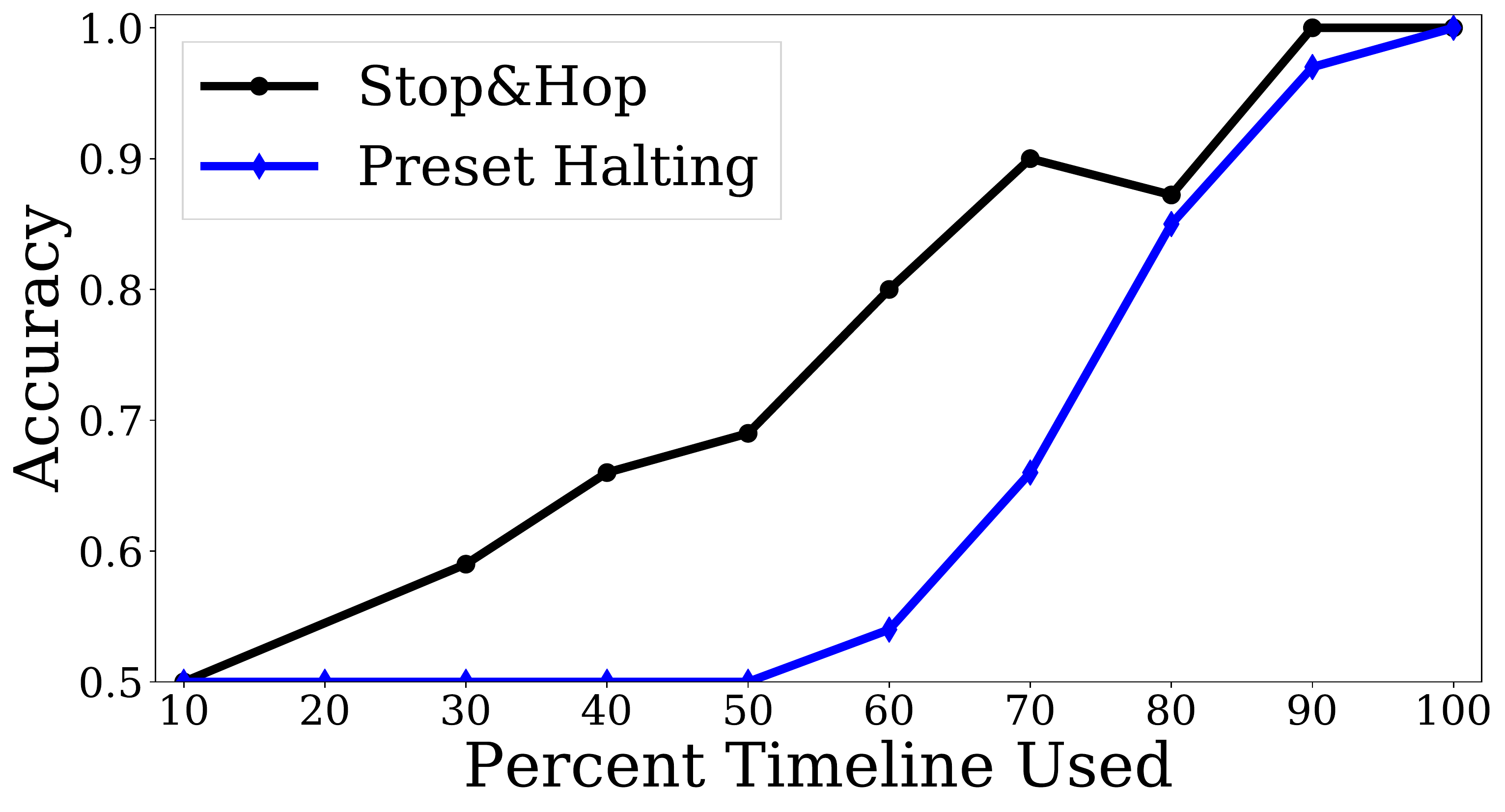}
        \vspace{-5mm}
        \caption{\textsc{Late} true signal times}
    \end{subfigure}
    % \begin{subfigure}{.24\linewidth}
    %     \includegraphics[width=\textwidth]{figures/synthetic_early_late.pdf}
    %     \vspace{-5mm}
    %     \caption{\textsc{BiModal} dataset}
    % \end{subfigure}
    \caption{Earliness vs. Accuracy on three synthetic datasets---\textsc{Uniform}, \textsc{Early}, and \textsc{Late}. The X axis denotes the \textit{average} percent of the timeline used as \method predicts one halting point per time series. The high black line indicates that \method succeeds to stop and classify irregular time series at effective times.
    }
    \label{fig:synthetic_results}
    % \vspace{-5mm}
\end{figure*}

\begin{figure*}[t]
    \centering
    \begin{subfigure}{.48\linewidth}
        \includegraphics[width=\textwidth]{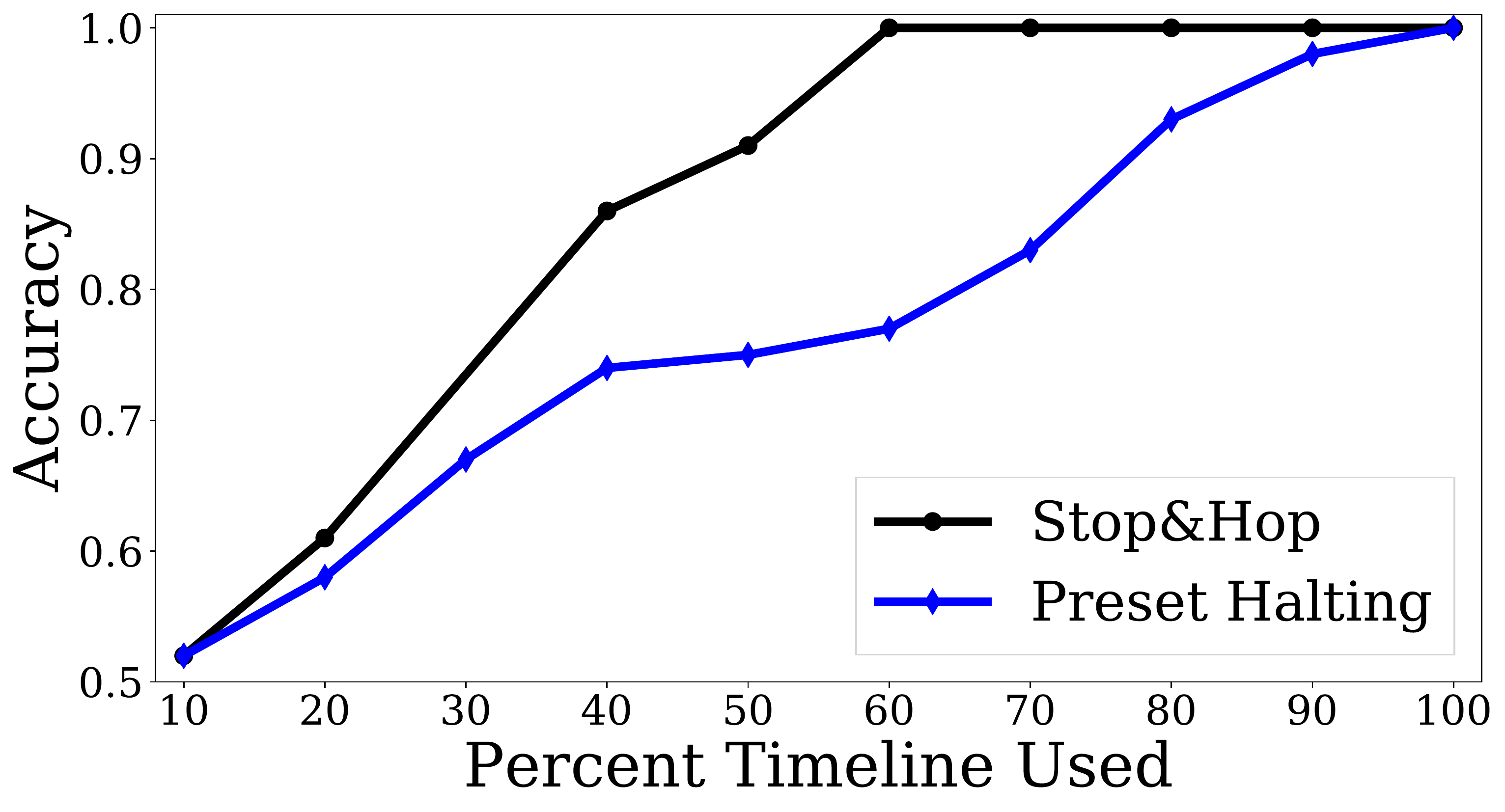}
        \vspace{-5mm}
        \caption{Earliness vs. Accuracy}
    \end{subfigure}
    \vspace{-2mm}
    \begin{subfigure}{0.48\linewidth}
        \includegraphics[width=\linewidth]{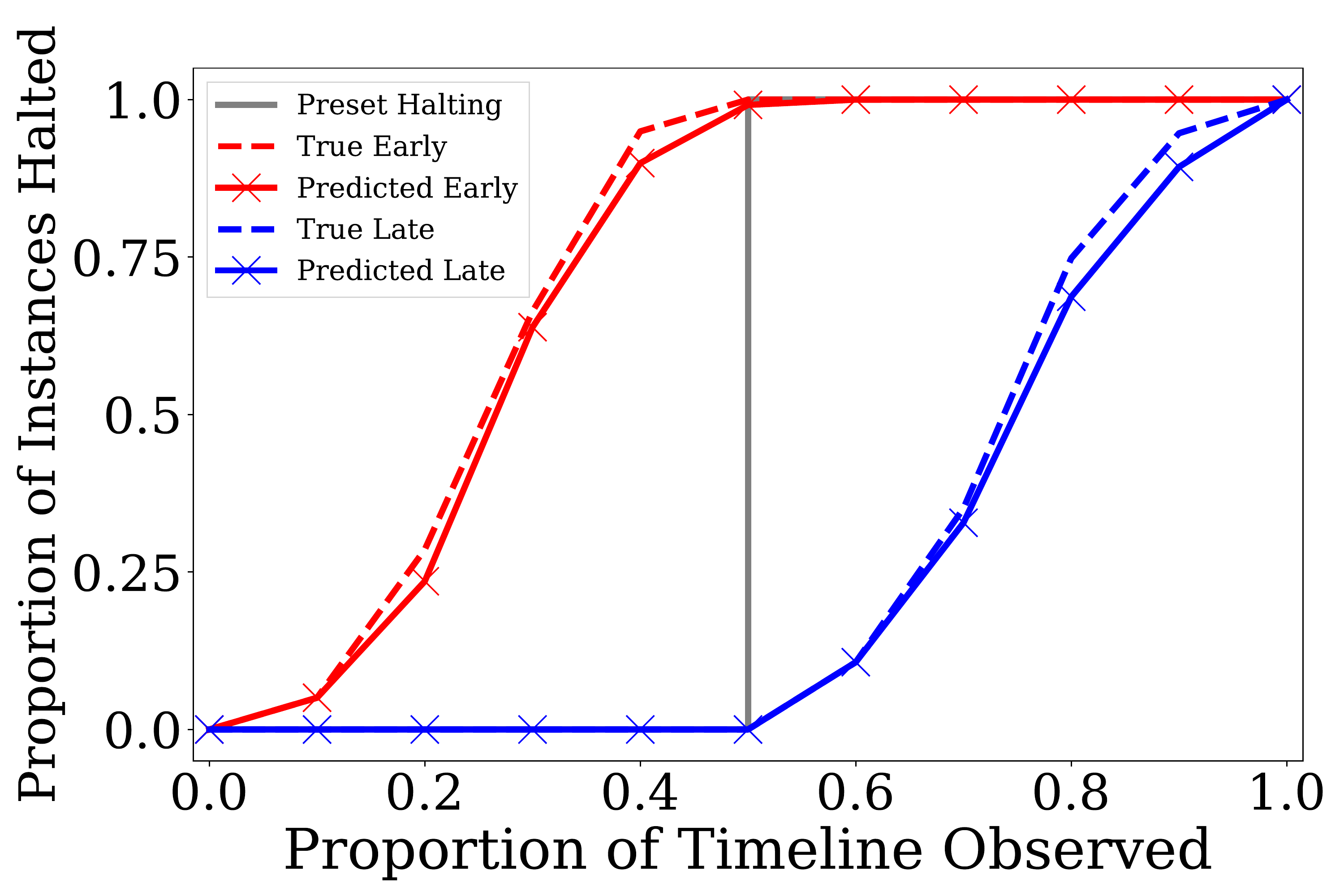}
        \caption{Cumulative halting distribution}\label{fig:cumulative}
    \end{subfigure}
    \caption{Results for \textsc{BiModal} dataset. In (a) \method outperforms Preset Halting. In (b), the solid and dashed lines match, indicating that \method is nearly optimal.}% halting times.}
    \label{fig:bimodal}
    \vspace{-5mm}
\end{figure*}

For the \textsc{PhysioNet} dataset, \method maintains improvement across all average halting times, converging with the comparisons once all time steps are observed---as expected.
Further, this AUC ($\sim$0.82) is nearly state-of-the-art for this dataset \cite{shukla2021multi}.
For the two \textsc{ExtraSensory} datasets, \method's improvements seem to come earlier in the timeline; at some point, AUC appears to saturate and \method and EARLIEST accurately classify all testing time series.
As we will show on our synthetic results (Figure \ref{fig:cumulative}), this happens when a method predicts good halting points.
We are also unsurprised by this saturation in general: there will always be cases where relevant windows of a time series are isolated \cite{zhu2021uncertainty}.
For these datasets, the optimal halting times are unknown, yet seem to be around 40 minutes for \textsc{Walking} and 60 minutes for \textsc{Running} on average.
Discovering good halting times retrospectively is an added benefit of a successful early classifier; it can even recover reasonable halting times for individual instances, depending on whether or not the classification was accurate.
In summary, \method consistently outperforms the alternatives on these three real-world datasets.

We also conduct a hyperparameter study for $\lambda$ for the \textsc{Running} dataset, shown in Figure \ref{fig:hyperparams}.
Our results indicate that $\lambda$ has strong control over the earliness--accuracy trade-off: As $\lambda$ increases, Halting Time and Accuracy steadily decrease.
Standard deviations are computed across five replications of the same experiment with different seeds.
Thus emphasizing earliness or accuracy is intuitive.
% Small values indicate that \method consistently learns a successful policy for these data.
% Further discussion is available in our supplementary materials.

To further ablate the impacts of hyperparemters on \method's success, we also consider the standard deviation $\sigma$ of the hop policy $\pi_\text{hop}$, which controls for how long to wait before trying to stop again. As $\sigma$ increases, so does the variance in chosen hop sizes. Interestingly, we find that \method's performance is largely robust to changing $\sigma$, as shown in Figure \ref{fig:std_hyper}. Still, varying $\sigma$ seems to have some impact and we recommend tuning it according to the task at hand. $\sigma$ also determines to what degree $\pi_\text{hop}$ directly controls the sampled hop size. As $\sigma$ grows, $\pi_\text{hop}$ has less direct control, which can improve exploration and regularize the chosen actions.

\subsection{\method finds the true halting times}
We next verify that \method indeed finds the true halting times by using use our four synthetic datasets where we know the halting times.
Our results are shown in Figures \ref{fig:synthetic_results} and \ref{fig:bimodal}, where we compare \method to a \textit{Preset Halting} baseline with the same prefix embedding approach as \method to isolate the effects of \textit{learning when to stop}.
The preset halting method stops at a set of predetermined halting times.
For example, a preset halting method that uses 50\% of the timeline stops all instances at time 0.5.
% To train \method, we tune $\lambda$, the \textit{wait penalty}, so that \method halts on average at 10 to 100\% of the timeline used and report the corresponding accuracy.
% We stop \textit{Preset Halting} at the same points for comparison.
% To demonstrate that the halting policy network learns an effective halting policy, we let it choose between stopping and hopping forwards with $\delta=0.1$.
% One $\tau$ is predicted per series.

Our experiments show that \method clearly achieves higher accuracy while using less of the timeline compared to preset halting times, as expected.
This is only possible if \method appropriately halts when it sees a signal and waits otherwise.
% Thus the halting policy network succeeds to learn when to Stop and when to Hop.
As the four synthetic datasets have different halting distributions, 
% clearly shown in the preset halting comparison's accuracy trend,
we see that \method succeeds to wait longer when signals are all later (Figure \ref{fig:synthetic_results}c) and stop earlier when signals are all earlier (Figure \ref{fig:synthetic_results}b).
Note that for the \textsc{Uniform} dataset, \method steadily increases to Accuracy of 1.0 (indicating 100\% accurate predictions), which makes sense because the true signals are distributed uniformly across the timeline.
For the \textsc{Early} dataset, where signals happen early in the timeline, AUC saturates early and Accuracy steeply increases early on.
For \textsc{Late}, Accuracy increases slowly early on and steeply later.
In each case, the Preset Halting baseline's Accuracy changes exactly as expected.

For the \textsc{BiModal} dataset---highlighted in Figure \ref{fig:bimodal}---we first observe that \method again makes early and accurate predictions, even when signals are distributed unevenly across the timeline.
Its curve is characterized by a steep increase, followed by a plateau, followed by a steep increase, and finally a plateau.
We expand this experiment and also show a snapshot of the halting distribution from the \textsc{BiModal} dataset from \method trained with $\lambda=3e^{-6}$ in Figure \ref{fig:cumulative}.
Each instance's predicted halting time is plotted against the proportion of the dataset with halting times earlier than a set of possible halting times, showing the cumulative halting distribution.
We color-code the early and late signals and find that \method matches the cumulative frequencies of the halting timings almost \textit{perfectly}.
% \method solves this task almost \textit{perfectly}, even without access to the true halting times. \wnote{the first part of this kind of repeats the last sentence of the previous paragraph}
As our method captures all positives exactly on time, matching the true cumulative functions without supervision, we postulate that \method can also learn other complex functions.
% \wnote{i think this is a super cool result, and I wonder if we could play it up a bit more. the results imply it's doing about as good as it could possibly do, right? }
In contrast, the preset halting comparison's halting distribution is a step function: All instances halt at the same time. This is not flexible enough to match the halting distributions of real datasets.

\section{Ethical Considerations}
Our work facilitates decision making given partial temporal information.
This lack of knowledge can naturally lead to misclassifications, which are more or less dangerous depending on the task. % and specifics of a given time series.
For example, incorrectly predicting \textit{Cancer} early may cause undue stress and financial burden to a patient.
Still, the cost of risking a false positive must be balanced with the cost of delaying predictions, which itself may have negative impacts.
Early classifiers do not \textit{suffer from} or \textit{introduce} this trade-off, and instead \textit{embrace reality}---the trade-off is real in practice, regardless of the algorithm.
Standard classifiers, for instance, ignore prediction timing, so they always pick one side of the trade-off.
Algorithms like \method crucially allow end-users to balance their own risk aversion. 

% \section{Limitations and Potential Negative Impacts}
% Our work facilitates decision making given partial temporal information.
% This can naturally lead to false positives, which are more or less dangerous depending on the task and specifics of a given time series.
% For example, an early \textit{Cancer} prediction may cause undue stress and financial burden to a patient if it turns out to be wrong.
% Still, the cost of risking a false positive must be balanced with the cost of delaying predictions, which itself may have negative impacts.
% We pose that non-early classifiers already incur this cost naturally as they do not consider the timing of a prediction for a given instance anyways, generating all of their predictions concurrently without learning \textit{when} to make a prediction.

% In our experiments, we treat hop option selection as a categorical prediction problem, letting our method choose how long to wait before trying to stop again from a \textit{set} of options.
% In principle, this limits the flexibility of the system since our method still works with any real-valued candidate halting times.
% However, such continuous sampling is notoriously challenging and tackling this problem is out of scope for this work.
% While this simplification of the learning process can lead to more robust policies, it may also bias our solution towards later predictions than are necessary if selected the hop options are too large.
% In future work, learning continuous values for hop sizes is an interesting and promising direction that can further advance solutions to the ECITS problem.

\section{Conclusion}
\label{sec:conclusions}
Our work introduces the open Early Classification of Irregular Time Series problem.
This is an important, challenging, and interesting problem that has yet to be considered by the early classification community.
%We kick off this direction by
We provide a general formulation for solving this problem, which we instantiate as a modular framework that serves as an effective first solution, named \method.
\method is a novel reinforcement learning-based continuous-time recurrent network that leverages irregularity in the inputs to inform classifications that are both early and accurate.
By using irregularity to inform \textit{when} to classify ongoing series, \method advances beyond the state-of-the-art for early classification.
Our \textit{irregularity-aware halting policy network}  chooses \textit{when} to stop, which allows more-flexible halting policies than recent alternatives and shortens the reinforcement learning trajectories, leading to more stable training.
% \wnote{``relative to other recent halting methods" or something? ``recent similar methods" might come across as our method being a small delta}
%% odd sentence below.
%% We verify that \method is indeed the state-of-the-art for our problem by using three real-world public datasets and four synthetic datasets.
% We  demonstrate   \method's effectiveness on three real-world public and four synthetic datasets.
We find that \method indeed halts at the earliest-possible times on all four  synthetic data sets
and consistently outperforms alternatives on 
all three  real datasets by making earlier and more-accurate predictions.

% This new problem is much more realistic than existing early classification settings.
% Our approach is a novel reinforcement learning-based continuous-time recurrent network that instantiates a modular and general framework to solve the ECITS problem.
% %, marrying state-of-the-art ITS representation learning with a reinforcement learning agent.
% Our model learns to use patterns in both the values and irregularity of ongoing ITS to predict when enough data has been collected to warrant classification \textit{with respect to the task at hand}.
% This is achieved without direct supervision on the halting times and our solution is naturally tunable between goals of \textit{accuracy} and \textit{earliness}.
% Using four synthetic datasets, we demonstrate that \method halts at the earliest possible times.
% We then validate \method on three real-world public datasets, demonstrating that our method outperforms existing methods in all settings, making early and accurate predictions. 
% \wnote{should we make this even stronger and say how StopHop is NEVER outperformed by the existing methods, at any stopping point? or leave that for the results section?}
% that are more accurate and earlier than those of state-of-the-art alternatives. 

% \wnote{should we reiterate how irregular time series is a more realistic assumption or the like here in the conclusion? to show not just that we solve this new problem, but a very realsitic/useful one?}

With this work, we also advocate for broadening the evaluation of machine predictions.
Our community often evaluates machine learning models using solely accuracy (or a similar measure).
However, in time-sensitive domains, accuracy is irrelevant if a model is used too late.
We pose that earliness, along with other highly-impactful directions, like fairness and explainability, should also be considered when developing machine learning systems. 
% Thinking along these lines will lead to  fruitful research.

% Our work is also a step forward in promoting the \textit{practicality} of machine predictions. % in a new light.
% Accuracy (or a similar measure) is often used alone to evaluate a model.
% However, in time-sensitive domains, a classifier that learns to make earlier predictions and trade off some accuracy is preferred.
% There has been a large amount of recent work in many important directions---such as fairness and explainability---and we hope our work will also encourage the community to pursue measures of success beyond pure accuracy.
% In particular, the \textit{timing} of a machine learning model's predictions can heavily impact their usefulness.
% \wnote{i like this last paragraph. it's neat to claim that the contribution isn't just a method specific to this problem, but inspiring people to think about different evaluations in general}

\section{Acknowledgements}
This research was supported in part by grants from U.S. Dept. of Education P200A150306, NSF-IIS 1910880 and NSF-OAC 2103832. We are grateful to the WPI DAISY lab for constructive feedback and support.

\balance
\bibliographystyle{ACM-Reference-Format}
\bibliography{main.bib}

\end{document}